%% file: main_arxiv.tex
\documentclass[letterpaper]{article}
\usepackage[preprint]{aaai2027_arxiv}
\input{preamble}
\title{EgoAfford: A Dataset and Benchmark for Role-Structured Egocentric Affordance Grounding}
\author {
    Xinyuan Guan\textsuperscript{\rm 1,\rm 2},
    Feifan Chen\textsuperscript{\rm 1},
    Xinyu Zhan\textsuperscript{\rm 1},
    Fu-Cheng Zhang\textsuperscript{\rm 2},
    Cewu Lu\textsuperscript{\rm 1,\rm 2,\rm 3},
    Lixin Yang\textsuperscript{\rm 1,\rm 2}\corresponding
}
\affiliations {
    \textsuperscript{\rm 1}Shanghai Jiao Tong University;\quad
    \textsuperscript{\rm 2}Shanghai Innovation Institute;\quad
    \textsuperscript{\rm 3}Noematrix Ltd\\
    \texttt{chapman\_guan@sjtu.edu.cn; siriusyang@sjtu.edu.cn}
}

\begin{document}
\maketitle
\raggedbottom
\input{fig/teaser}
\input{sec/0_abstract}
\input{sec/1_introduction}
\input{sec/2_related_work}
\input{sec/3_task_protocol}
\input{sec/4_dataset}
\input{sec/5_baselines}
\input{sec/6_experiments}

\input{sec/7_manipulation}
\FloatBarrier
\input{sec/8_discussion}
\clearpage
{\small\bibliography{main}}
\input{sec/X_appendix}
\end{document}

%% file: preamble.tex
\usepackage[hyphens]{url}
\usepackage{graphicx}
\usepackage{natbib}
\usepackage{caption}
\usepackage{subcaption}
\usepackage{amsmath,amssymb}
\usepackage{booktabs,tabularx,array}
\newcolumntype{Y}{>{\raggedright\arraybackslash}X}
\usepackage{cuted}
\usepackage{float,needspace}
\usepackage[shortlabels,inline]{enumitem}
\usepackage{xspace}
\definecolor{cvprblue}{rgb}{0.21,0.49,0.74}
\usepackage[pagebackref,breaklinks,colorlinks,allcolors=cvprblue,hypertexnames=false]{hyperref}
\usepackage[capitalize]{cleveref}
\crefname{section}{Sec.}{Secs.}
\Crefname{section}{Section}{Sections}
\crefname{table}{Tab.}{Tabs.}
\Crefname{table}{Table}{Tables}
\hypersetup{
  pdftitle={EgoAfford: A Dataset and Benchmark for Role-Structured Egocentric Affordance Grounding},
  pdfauthor={Xinyuan Guan, Feifan Chen, Xinyu Zhan, Fu-Cheng Zhang, Cewu Lu, Lixin Yang}
}

\newcommand{\figtabnote}{\fontsize{8}{9.5}\selectfont}
\crefname{appfigure}{Appendix Fig.}{Appendix Figs.}
\Crefname{appfigure}{Appendix Figure}{Appendix Figures}
\crefname{apptable}{Appendix Tab.}{Appendix Tabs.}
\Crefname{apptable}{Appendix Table}{Appendix Tables}

%% file: fig/teaser.tex
\begin{strip}
    \centering
    \includegraphics[width=\linewidth]{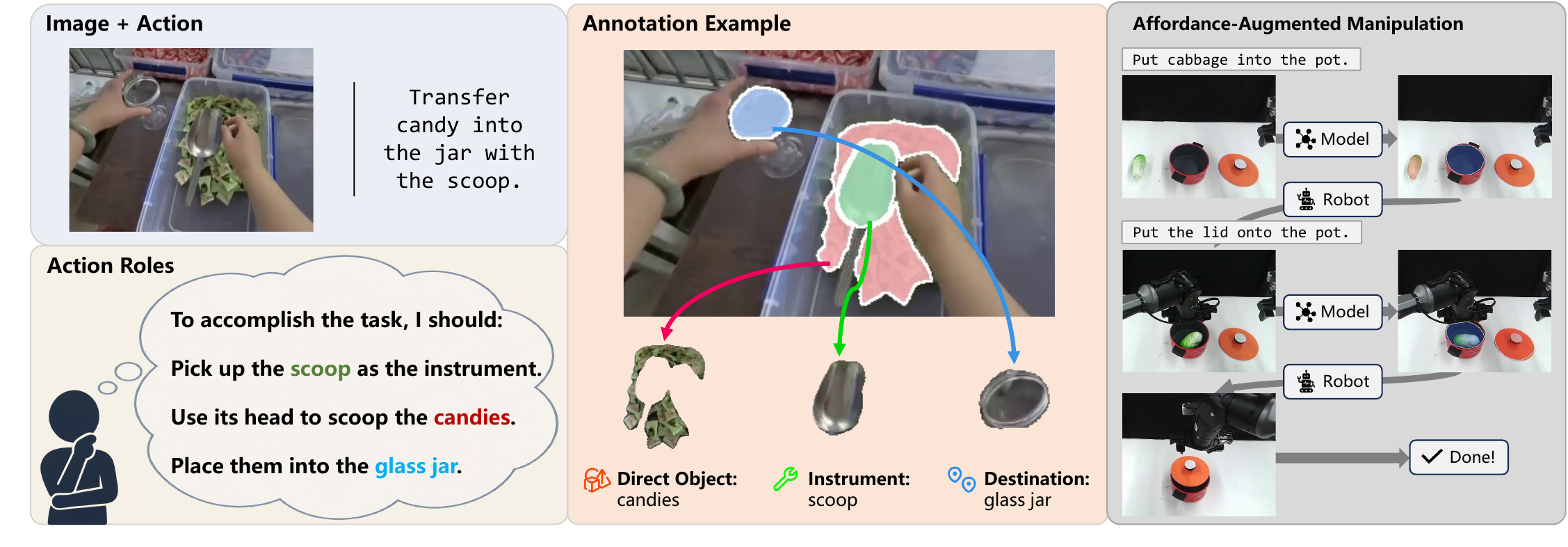}
    \captionsetup{hypcap=false}
    \captionof{figure}{\textbf{Role-structured grounding and manipulation with EgoAfford.} Left: an egocentric observation and the instruction to transfer candy into a jar with a scoop introduce the direct object, instrument, and destination roles. Middle: the corresponding annotations identify the candies, the scoop's working region, and the jar opening, with enlarged views of these regions. Right: role masks guide an agentic robot through cabbage placement and lid closure, with grounding refreshed for each step. Red, green, and blue denote O, I, and D.}
    \label{fig:teaser}
    \captionsetup{hypcap=true}
\end{strip}

%% file: sec/0_abstract.tex
\begin{abstract}
Part-level affordance grounding localizes functional object regions associated with actions. Manipulation also requires distinguishing the roles of multiple participating objects. We introduce EgoAfford, a dataset and benchmark for action-conditioned role-structured grounding from egocentric images. Given a current image and an action instruction, a model segments action-relevant regions for the direct object, instrument, and destination, with empty masks for absent roles. EgoAfford combines real observations drawn from egocentric videos with synthetic observations organized around task states under a shared annotation schema. Training uses 26,250 real and 13,706 synthetic image--action pairs, with a human-reviewed real test set. Evaluations reveal challenges in role assignment and fine-grained localization. EgoAfford training improves MLLM grounding; an 8B-backbone model achieves real-test cIoU comparable to Astra--SAM2 (61.12\% vs. 59.78\%). We further conduct two manipulation studies in LIBERO-Goal and on a real robot, showing that role-part masks improve learned-policy execution and agentic task completion. Project page: \url{https://egoafford.github.io}.
\end{abstract}

%% file: sec/1_introduction.tex
\section{Introduction}
\label{sec:intro}
Affordance describes the action possibilities an environment offers an agent, bridging perception and interaction~\cite{gibson2014ecological}. In object manipulation, affordances are often localized to action-relevant functional parts, such as a blade for cutting or a spout for pouring. Recent benchmarks~\cite{wan2025instructpart,liu2026affordance2action} extend this view to task-conditioned part segmentation, identifying functional regions from action instructions.

Manipulation often involves several objects that contribute differently to the same action. We represent each action with three components: the direct object being manipulated, the instrument used to act on it, and the destination receiving the object or transferred material. Grounding these roles at part level binds each participating object to its action-relevant region and its role in the interaction. For example, pouring requires identifying the source container's pouring region and the receiving region of the destination, whereas tool use also requires localizing the instrument's working part. \Cref{fig:teaser} gives an overview of this role-structured representation.

Learning this representation calls for observations that bind action-specific roles to participating objects and their functional regions. The same object category can take different roles across actions, and the relevant region depends on how the object is used. A benchmark for role-structured part grounding should therefore combine diverse objects and interactions with consistent annotations of roles, functional regions, and absent components.

To provide this supervision, we build EgoAfford, a dataset and benchmark for role-structured part grounding. The model receives a current egocentric image and an action instruction and predicts masks for the direct object, instrument, and destination. EgoAfford combines real observations annotated from EgoStandard videos~\cite{lightwheel2026egosuite} with synthetic observations organized around task states under a common role-based schema. The training set contains \textbf{26,250 real and 13,706 synthetic image--action samples.} The real test set contains \textbf{2,215 human-reviewed examples with pixel-level role-part masks}, complemented by 488 synthetic test examples.

Using this resource, we benchmark representative referring, reasoning, and affordance-specialist segmentation models under zero-shot prompting and after task-adapted training on EgoAfford. \textbf{Training on EgoAfford substantially improves grounding on both test sets.} After training on EgoAfford, Sa2VA with an 8B backbone achieves \textbf{61.12\% cIoU on the real test set, comparable to Astra--SAM2 at 59.78\%}. The evaluation examines data scaling and source composition and diagnoses action conditioning and role-wise localization.

We further demonstrate the use of predicted role-part masks in a controlled manipulation study, with a domain-adapted grounding model conditioning an ACT-derived policy. An agentic real-robot study complements this learned-policy evaluation, testing role conditioning with GPT-6-luna (Luna) and GPT-6-astra (Astra) executors across three manipulation tasks. Role-part masks raise policy execution success \textbf{from 80.4\% to 88.2\%} and Luna's mean real-robot task-completion score \textbf{from 35.0\% to 58.3\%}.

Our contributions are threefold:
\begin{itemize}
    \item We formulate action-conditioned role-structured part grounding through masks for the direct object, instrument, and destination.
    \item We present EgoAfford, combining real and synthetic egocentric observations with consistent role-part annotations and a human-reviewed real test set with pixel-level masks.
    \item We establish task-adapted baselines that substantially benefit from EgoAfford supervision, alongside systematic evaluations of data scaling, source composition, role-wise grounding, and downstream utility in policy-based and agentic manipulation.
\end{itemize}

%% file: sec/2_related_work.tex
\section{Related work}
\label{sec:related}

\subsection{Affordance understanding}
\label{sec:related-1}
Affordance grounding has developed from functional-category segmentation~\cite{myers2015affordance,nguyen2017object,do2018affordancenet} toward learning from natural images, human interactions, and VLM knowledge~\cite{zhai2022one,luo2022learning,qian2024affordancellm}.
Related studies examine object attributes and contextual action possibilities~\cite{li2023beyond,chuang2018learning}, as well as functional regions on 3D objects and in scenes~\cite{deng20213d,li2024laso,yu2025seqafford,he2026task,wu2026affordany}. Hand--object resources connect affordances with interaction geometry and motion~\cite{Liu_2022_CVPR,heidinger20252handedafforder}. In particular, OakInk and OakInk2 link object functionality to manipulation intents and primitive tasks~\cite{yang2022oakink,zhan2024oakink2}.

Language-conditioned resources associate instructions with functional masks~\cite{wan2025instructpart,wang2026affordance,wu2025ragnet}, interaction points~\cite{ma2025glover++}, or box and point targets~\cite{hao2025roboafford++}.
Related work studies contact transfer and point prediction~\cite{ju2024robo,yuan2024robopoint}, manipulation knowledge~\cite{huang2024manipvqa}, and egocentric intention grounding~\cite{sun2025visual}.
WorldAfford and A2A-Bench extend task-conditioned grounding to multiple objects and functional regions~\cite{chen2024worldafford,liu2026affordance2action}, while AFUN pairs functional masks with post-contact motion~\cite{wang2026afun}. EgoAfford binds each action's direct-object, instrument, or destination role to its action-relevant region, giving real and synthetic egocentric observations a consistent role-part schema with explicit absent components.

\subsection{Language-guided segmentation}
\label{sec:related-2}
Referring expression comprehension and segmentation ground language queries in image regions~\cite{hu2016segmentation,yu2018mattnet,yang2022lavt,wang2022cris}. MLLM-based methods extend this interface to reasoning segmentation~\cite{lai2024lisa}, multiple or absent targets~\cite{ren2024pixellm,xia2024gsva}, and grounded language generation~\cite{rasheed2024glamm,zhang2024omg}.

Recent models provide different mask-generation interfaces: Sa2VA couples multimodal language understanding with SAM2~\cite{yuan2025sa2va,ravi2024sam2}, LENS and UniPixel develop segmentation-oriented reasoning~\cite{zhu2026lens,liu2026unipixel}, and the SAMTok mask tokenizer represents masks with discrete tokens~\cite{zhou2026samtok}. We benchmark and adapt models built on these interfaces for action-conditioned role-structured part grounding, requiring aligned masks for multiple roles and empty masks for absent components.

%% file: sec/3_task_protocol.tex
\section{Task and evaluation}
\label{sec:task}

\subsection{Problem formulation}
\label{sec:task-1}
Given an egocentric image $I$ and a natural-language action instruction $T$, a model predicts an ordered tuple of masks $(\hat M_o,\hat M_i,\hat M_d)$ for the direct object, instrument, and destination of the action. Unhatted masks denote ground truth. These roles describe each participant's contribution:

\begin{itemize}
    \item \textbf{Direct object} ($M_o$): the object being manipulated, whether by bare hands or through an instrument.
    \item \textbf{Instrument} ($M_i$): the object held in hand to act on the direct object; absent when the action is performed with bare hands.
    \item \textbf{Destination} ($M_d$): the target location or object receiving a transferred object or material; absent when no transfer occurs.
\end{itemize}

Each non-empty mask delineates the action-relevant region of its component, such as the blade of a knife for cutting or the spout of a pot for pouring. For simple pick-and-place actions, the annotation can cover the whole object. An absent component corresponds to an all-zero mask.

Two additional rules specify how materials are represented. When the manipulated material resides in a container (e.g., water in a pot), the container is annotated as the direct object. A loaded instrument carrying material (e.g., a spoonful of syrup) is annotated as the instrument together with its carried material; the direct-object role may then be absent.
See \cref{sec:supp-annotation-rules} for examples of these material-handling conventions.

\subsection{Evaluation metrics}
\label{sec:task-2}
We report cIoU as the primary metric and gIoU as a complementary measure. gIoU averages the per-mask IoU over all evaluation samples and the three roles, whereas cIoU is the ratio between the globally accumulated intersection and union. Each prediction is evaluated against the annotation for the given action, with masks aligned in the fixed direct-object, instrument, and destination order.
If the ground truth is empty, the per-mask IoU is $1$ when the prediction is also empty and $0$ otherwise. For cIoU, an empty--empty pair contributes nothing to either accumulator, while a false positive contributes to the union. Thus, gIoU accounts for sample-level localization and recognition of absent components, whereas cIoU reflects overall pixel-level segmentation accuracy.

We also report role-wise non-empty mIoU (NE-mIoU), the mean per-sample IoU over cases with non-empty ground truth for the corresponding role. Empty predictions in these cases receive zero IoU; cases with empty ground truth are excluded. This measures grounding of present roles without credit for empty--empty pairs. Role-wise presence F1 treats a non-empty mask as a positive prediction; macro-F1 averages the three role-wise F1 scores.

%% file: sec/4_dataset.tex
\section{EgoAfford dataset}
\label{sec:dataset}

\subsection{Dataset overview}
\label{sec:dataset-1}
Learning role-structured part grounding requires image--action observations with consistent annotations of participating objects and their functional regions. We construct EgoAfford from two sources: real egocentric observations drawn from recorded interactions, and synthetic images organized around task states. Both follow the action-conditioned annotation schema in \cref{sec:task-1}.

EgoStandard~\cite{lightwheel2026egosuite} provides egocentric videos with subtask segments that can be converted into image--action samples. After filtering and annotation, we retain 26,250 real training samples, 778 validation samples, and 2,215 test samples from non-overlapping episodes. Real observations provide object appearances and interaction states encountered during actual manipulation. Synthetic generation complements this source by explicitly organizing object configurations and pre-action states. We retain 13,706 synthetic training samples and 488 samples for test. Synthetic splits share no generated scenes or task instances. The combined training set contains 39,956 samples.
Under a shared, manually reviewed mapping (\cref{tab:dataset-stats}), synthetic training adds 131 action and 636 object concepts absent from real training; real training adds 37 and 244 absent from synthetic training. Their union covers 353 actions and 1,322 objects.
Further statistics appear in \cref{sec:supp-statistics}.

\begin{table}[t]
    \centering
    \caption{\textbf{Sample counts and confirmed semantic concepts.} Concept counts use a shared, manually reviewed mapping.}
    \label{tab:dataset-stats}
    \small
    \setlength{\tabcolsep}{4pt}
    \begin{tabularx}{\linewidth}{@{}Yrrrr@{}}
        \toprule
        Split / source & Samples & Actions & Objects & Unique A/O \\
        \midrule
        Real train & 26,250 & 222 & 686 & 37/244 \\
        Synthetic train & 13,706 & 316 & 1,078 & 131/636 \\
        Mixed train & 39,956 & 353 & 1,322 & -- \\
        \midrule
        Real test & 2,215 & 127 & 308 & 88/256 \\
        Synthetic test & 488 & 61 & 162 & 22/110 \\
        \bottomrule
    \end{tabularx}
    \par\smallskip
    \begin{minipage}{\linewidth}\figtabnote
        Mixed concept counts are vocabulary unions. Unique A/O counts action/object concepts absent from the other source within the same split (train or test); it is not applicable to Mixed.
    \end{minipage}
\end{table}

\begin{figure*}[t]
    \centering
    \includegraphics[width=\linewidth]{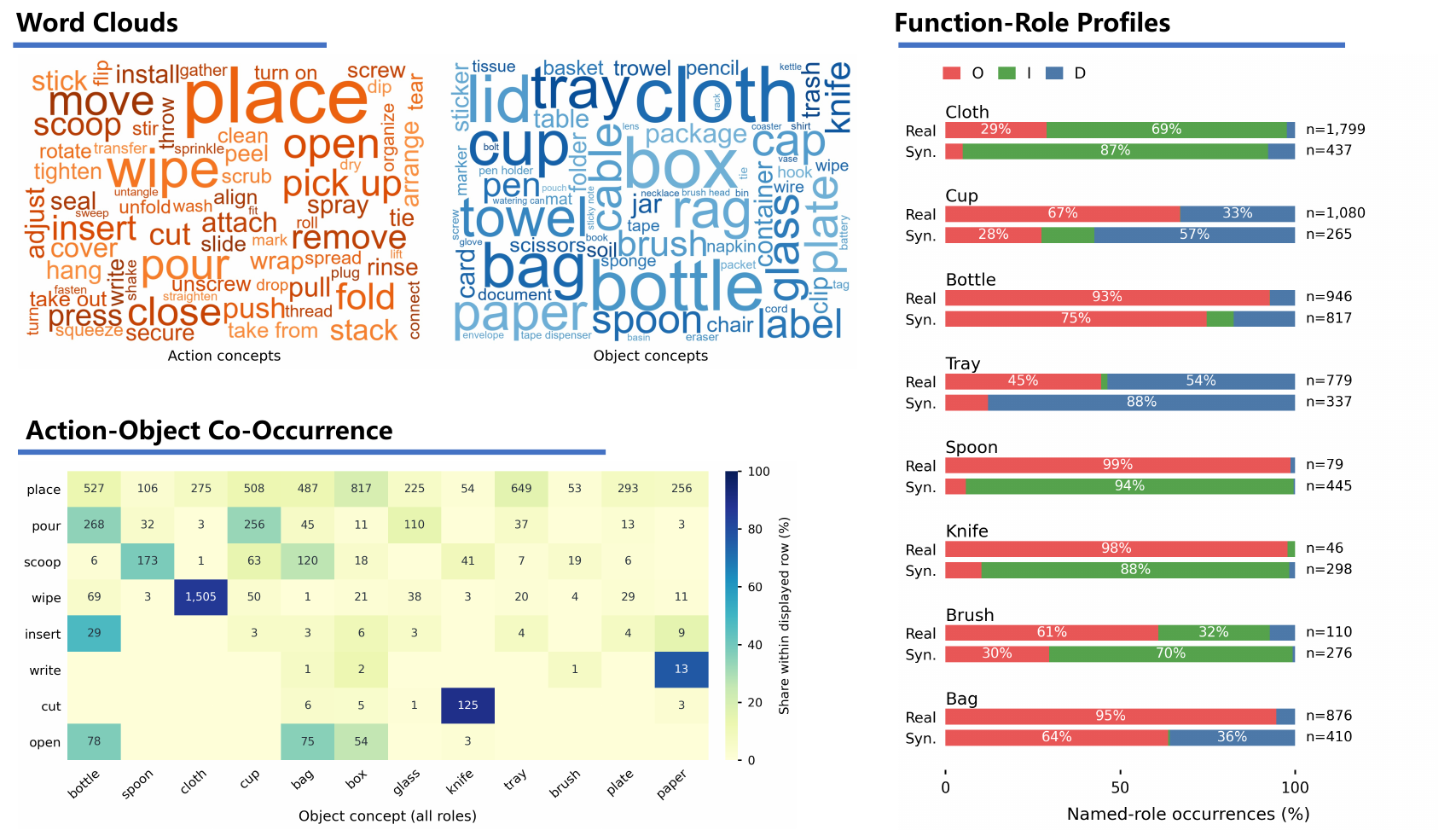}
    \caption{\textbf{Training semantics and role profiles.} Top left: word size reflects pooled record counts. Bottom left: action--object counts across all roles; colors indicate shares within each displayed row. Right: source-specific O/I/D profiles, normalized by named-role occurrences ($n$).}
    \label{fig:dataset-coverage}
\end{figure*}

\Cref{fig:dataset-coverage} shows action--object pairings and role distributions. Actions such as \emph{place} span many object concepts; cups and trays serve as both direct objects and destinations. Spoons and knives are predominantly direct objects in real observations and instruments in synthetic observations, highlighting the need to infer roles from the image and action beyond object identity.

Multiple roles are present in 73.9\% of real and 74.5\% of synthetic training samples, as determined by non-empty masks. Instruments occur in 18.2\% and 30.9\% of samples, respectively. Vocabulary overlap is summarized in \cref{tab:dataset-stats}.

Instrument regions have median areas of 0.42\% and 0.34\% of the image in real and synthetic training, respectively; 76.6\% and 77.1\% occupy at most 1\% of the image. These small functional regions make fine spatial localization a recurring requirement. Full area distributions appear in \cref{fig:supp-role-scales}.

\subsection{Dataset construction}
\label{sec:dataset-2}
\begin{figure*}[t]
    \centering
    \includegraphics[width=\linewidth]{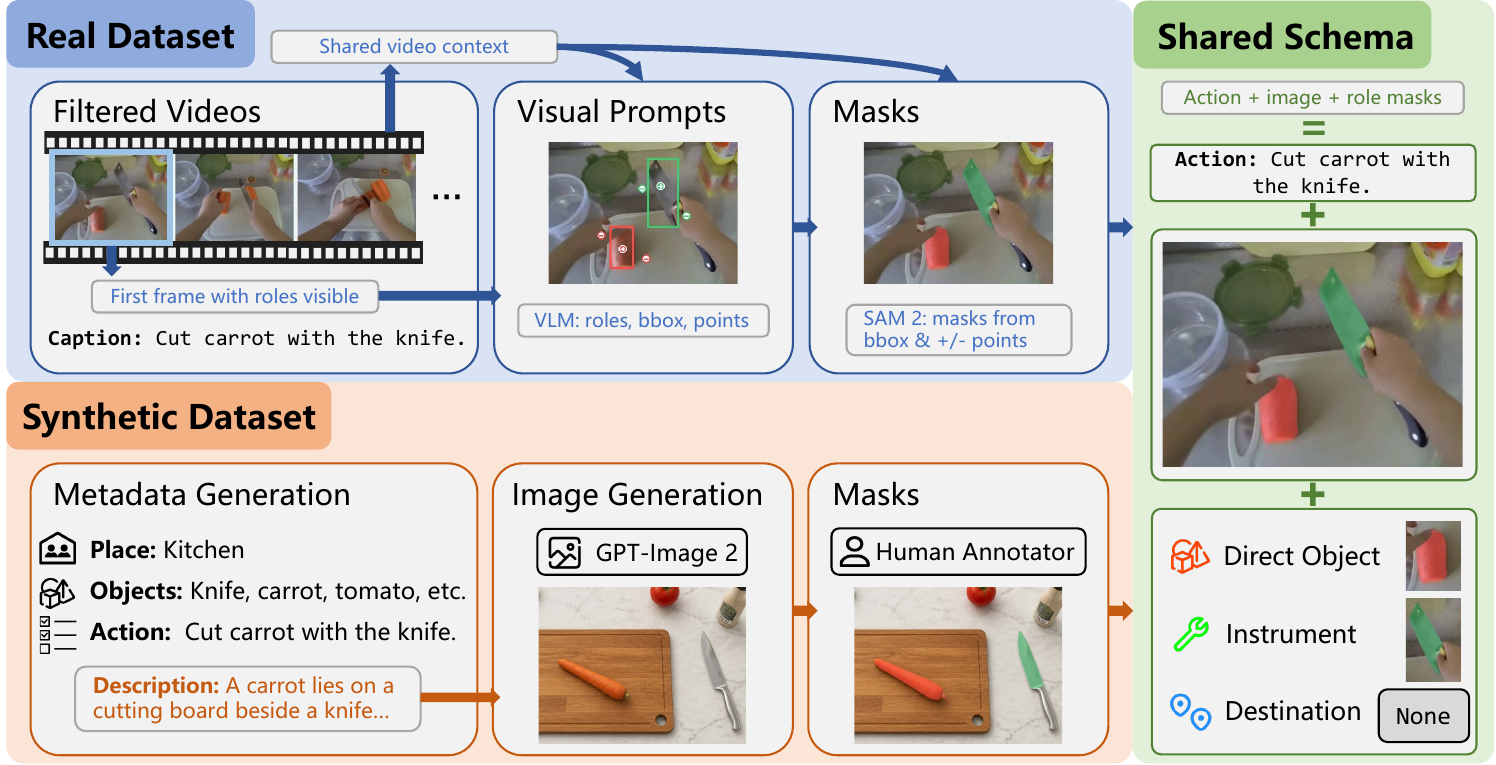}
    \caption{\textbf{Two data-construction paths with a shared output schema.} Top: filtered real videos provide context for selecting an early frame with visible action participants; a VLM supplies role-specific boxes and points to prompt SAM2. Bottom: task and scene metadata guide image generation, followed by manual role-part annotation. Right: both paths yield an image, an action, and O/I/D masks. The cutting example marks the carrot as the direct object and the knife's working region as the instrument, with no destination. Video context is used during annotation; benchmark inputs are the selected image and action.}
    \label{fig:construction}
\end{figure*}

The two sources share the same annotation goal but provide different contextual evidence. Video supports action verification and frame selection for real observations; synthetic image--action pairs receive manual role and region annotation. \Cref{fig:construction} summarizes these paths.

\paragraph{Real observations.}
We construct real image--action samples from the native subtask segments in EgoStandard. Metadata and caption-based filtering first remove unsuitable candidates. A VLM examines sparsely sampled video frames to verify the interaction and selects an early frame in which the relevant objects are visible. It identifies the participating roles and provides bounding boxes and positive and negative points, which jointly prompt SAM2~\cite{ravi2024sam2} to produce role-specific masks. After segmentation, a VLM checks and refines the action instruction and role names using the mask-overlay image and video context.
Training and test candidates use this video-assisted annotation framework; test candidates also undergo human review. Of 2,330 test candidates, 2,022 are accepted, 193 are retained after manual refinement of their action captions, role annotations, and masks, and 115 are rejected, yielding 2,215 image--action samples.

\paragraph{Synthetic observations.}
We organize synthetic generation around tasks and their successive scene states. LLMs generate task and object configurations and describe the scene before each action, yielding a scene description paired with an action instruction. We render these descriptions with GPT-Image-2. Generated images undergo manual screening; those with state inconsistencies are regenerated from corrected descriptions or discarded. For these static image--action samples, human annotators identify the direct object, instrument, and destination and delineate each action-relevant region using an interactive SAM2 interface. Absent roles receive empty masks.
Material-handling examples and annotation review are described in \cref{sec:supp-dataset}.

\subsection{Dataset comparison}
\label{sec:dataset-3}
\Cref{tab:dataset-comparison} compares related affordance datasets. EgoAfford binds action-relevant regions to direct-object, instrument, and destination roles, including their absence. Its training vocabulary spans 353 action and 1,322 object concepts, providing broad semantic coverage for role-structured part grounding.

\begin{table*}[t]
    \centering
    \caption{\textbf{Comparison of mask-annotated 2D affordance datasets.} Roles label action participants; concept counts follow the authors' taxonomies.}
    \label{tab:dataset-comparison}
    \small
    \setlength{\tabcolsep}{3pt}
    \begin{tabular*}{\textwidth}{@{\extracolsep{\fill}}llccrr@{}}
        \toprule
        Dataset & Query & Roles & Origin & \#Objects & \#Actions \\
        \midrule
        PADv2~\cite{zhai2022one} & Support image & None & R & 103 & 39$^*$ \\
        HANDAL~\cite{guo2023handal} & -- & None & R & 17 & --$^\dagger$ \\
        InstructPart / ReasonAff$^\ddagger$~\cite{wan2025instructpart,wang2026affordance} & Ref.+Task & None & R & 48 & 37 \\
        RAGNet~\cite{wu2025ragnet} & Ref.+Task & None & R+S & 180 & --$^\dagger$ \\
        EgoAfford & Task & O/I/D & R+S & 1,322 & 353 \\
        \bottomrule
    \end{tabular*}
    \par\smallskip
    \begin{minipage}{\textwidth}\figtabnote
        Ref.: object/part reference; Task: action/goal/functional intent; R/S: real/synthetic.
        $^*$PADv2 counts affordance categories, not actions.
        $^\dagger$Primarily grasp-region grounding; no action count reported.
        $^\ddagger$Shared visual annotations; counts from InstructPart. EgoAfford counts its combined training vocabulary.
    \end{minipage}
\end{table*}

%% file: sec/5_baselines.tex
\section{Baseline models}
\label{sec:baselines}

\subsection{Evaluated models}
\label{sec:baselines-1}
To assess transfer to role-structured part grounding and the benefit of EgoAfford supervision, we evaluate seven segmentation models. They comprise four general referring or reasoning models, Sa2VA~\cite{yuan2025sa2va}, UniPixel~\cite{liu2026unipixel}, LENS~\cite{zhu2026lens}, and Qwen3-VL-8B-SAMTok~\cite{zhou2026samtok}, and three affordance-specialist models, AffordanceVLM from RAGNet~\cite{wu2025ragnet}, Affordance-R1~\cite{wang2026affordance}, and AFUN~\cite{wang2026afun}. For each model, we compare its public checkpoint under zero-shot prompting with a task-adapted model trained on EgoAfford. Their different output interfaces test whether the same role-part supervision transfers across model families. The shared evaluation protocol and model variants are given in \cref{sec:experiments-1,tab:main-benchmark}.

We also evaluate an off-the-shelf VLM--SAM2 pipeline. Given the image and action instruction, a commercial VLM identifies the participating roles and predicts bounding boxes and point prompts, which SAM2~\cite{ravi2024sam2} converts into masks. This separates semantic grounding from mask generation and provides a reference for performance obtainable with existing tools.

\subsection{Model adaptation}
\label{sec:baselines-2}
Our task requires masks with explicit direct-object, instrument, and destination identities. For LENS, we triple the width of the final projection producing SAM2 visual prompts and split its output into three role prompts. Sa2VA and UniPixel receive role-indexed segmentation-token supervision. Qwen3-VL--SAMTok, which equips Qwen3-VL with the SAMTok codec, generates role-indexed mask codes. AffordanceVLM learns role-labeled \texttt{[AFF]} markers and masks; Affordance-R1 learns role-wise boxes and points for SAM2; AFUN uses role-indexed queries with its native SAM3 segmentation path.

Beyond these interface adaptations, we consider compatible task-rule prompts, auxiliary role-name and box targets, and initialization choices. For segmentation-token interfaces, \emph{emit} retains a token with an all-zero target for an absent role, while \emph{omit} removes that token; evaluation restores omitted roles as empty masks. We select each model's recipe using validation cIoU. Full selection comparisons, prompts, trainable components, and optimization settings are given in \cref{sec:supp-training}.

%% file: sec/6_experiments.tex
\section{Experiments}
\label{sec:experiments}

\subsection{Experimental setup}
\label{sec:experiments-1}
We use the union of the real and synthetic training sets for the main benchmark, comprising 39,956 image--action pairs. Checkpoints are selected by cIoU on the fixed 778-sample real validation set from EgoStandard and evaluated on the real test set (2,215 samples) and the synthetic test set (488). The data-scale and composition studies in \cref{sec:data-composition} use their specified training subsets and recipes.

We select each model's adaptation configuration on a shared mixed-data subset of 10,026 samples using validation cIoU, then train the selected configuration on the full training set. Candidate configurations, selected recipes, and model-specific optimization settings are detailed in \cref{sec:supp-training}.

For the four general models, joint prompting to zero-shot official checkpoints rarely yields outputs that satisfy the role-to-mask protocol. We therefore use three independent role-specific zero-shot queries with official checkpoints and assemble their masks in fixed O/I/D order. Trained models use their adapted interfaces. VLM--SAM2 uses GPT-6.1-sol or GPT-6-astra. Inference details and the joint-output comparison are provided in \cref{sec:supp-inference}.

\subsection{Benchmark results}
\label{sec:experiments-2}

\begin{table*}[t]
    \centering
    \caption{\textbf{EgoAfford results (cIoU, \%).} (a) Official zero-shot references (Before), selected task-adapted training (After), and gains ($\Delta$, points). (b) Best trained baseline versus VLM--SAM2 pipelines. Tests contain 2,215 real/488 synthetic samples. Bold/underline mark best/second-best trained scores in (a) and best/second-best scores in (b); gains are bold.}
    \label{tab:main-benchmark}
    \small
    \setlength{\tabcolsep}{5pt}
    \textit{(a) Learning from EgoAfford}\par\smallskip
    \begin{tabular*}{\linewidth}{@{\extracolsep{\fill}}lrrrrrr@{}}
        \toprule
        & \multicolumn{3}{c}{Real} & \multicolumn{3}{c}{Synthetic} \\
        \cmidrule(lr){2-4}\cmidrule(lr){5-7}
        Model & Before & After & $\Delta$ & Before & After & $\Delta$ \\
        \midrule
        \multicolumn{7}{l}{\textit{General referring and reasoning segmentation models}} \\
        \cmidrule(lr){1-7}
        LENS-3B & 15.28 & 56.43 & \textbf{+41.15} & 19.33 & 60.72 & \textbf{+41.39} \\
        UniPixel-7B & 21.02 & 54.65 & \textbf{+33.63} & 31.97 & 59.86 & \textbf{+27.89} \\
        Sa2VA-InternVL3-8B & 23.05 & \textbf{61.12} & \textbf{+38.07} & 34.53 & \textbf{67.16} & \textbf{+32.63} \\
        Qwen3-VL-8B-SAMTok & 27.48 & 58.79 & \textbf{+31.31} & 41.63 & 64.27 & \textbf{+22.64} \\
        \midrule
        \multicolumn{7}{l}{\textit{Affordance-specialist models}} \\
        \cmidrule(lr){1-7}
        AffordanceVLM-7B & 8.86 & 55.86 & \textbf{+47.00} & 8.85 & 59.04 & \textbf{+50.19} \\
        Affordance-R1-7B + SAM2 & 23.27 & 57.24 & \textbf{+33.97} & 30.66 & \underline{66.25} & \textbf{+35.59} \\
        AFUN-8B + SAM3 & 17.22 & \underline{58.81} & \textbf{+41.59} & 21.76 & 62.41 & \textbf{+40.65} \\
        \bottomrule
    \end{tabular*}
    \par\medskip
    \textit{(b) Best trained model and off-the-shelf pipelines}\par\smallskip
    \begin{tabularx}{0.67\linewidth}{@{}Yrr@{}}
        \toprule
        Model / pipeline & Real & Synthetic \\
        \midrule
        Sa2VA-8B, trained on EgoAfford & \textbf{61.12} & 67.16 \\
        GPT-6.1-sol + SAM2 & 59.00 & \textbf{71.22} \\
        GPT-6-astra + SAM2 & \underline{59.78} & \underline{70.87} \\
        \bottomrule
    \end{tabularx}
\end{table*}
The paired comparisons in \cref{tab:main-benchmark}(a) show substantial gains from EgoAfford training and task adaptation on both test sets across all seven models. On the real test set, our primary comparison set, LENS, UniPixel, Sa2VA, and Qwen3-VL--SAMTok improve cIoU by 41.15, 33.63, 38.07, and 31.31 percentage points, respectively; AffordanceVLM, Affordance-R1, and AFUN improve by 47.00, 33.97, and 41.59 points. Sa2VA achieves the highest trained-model cIoU on both test sets. These gains show the value of EgoAfford supervision across general segmentation and affordance-specialist models. Zero-shot evaluation uses separate role queries, whereas training also adapts the joint output interface.

\Cref{tab:main-benchmark}(b) compares the strongest trained baseline with the two tool pipelines. Trained on EgoAfford, Sa2VA with an 8B backbone reaches 61.12\% real-test cIoU, comparable to Sol--SAM2 and Astra--SAM2 at 59.00\% and 59.78\%, respectively. The rankings differ across domains: stronger synthetic-test scores do not necessarily translate to stronger grounding in real scenes. Complementary gIoU results are provided in \cref{tab:supp-main-giou}.

\begin{figure*}[t]
    \centering
    \includegraphics[width=\linewidth]{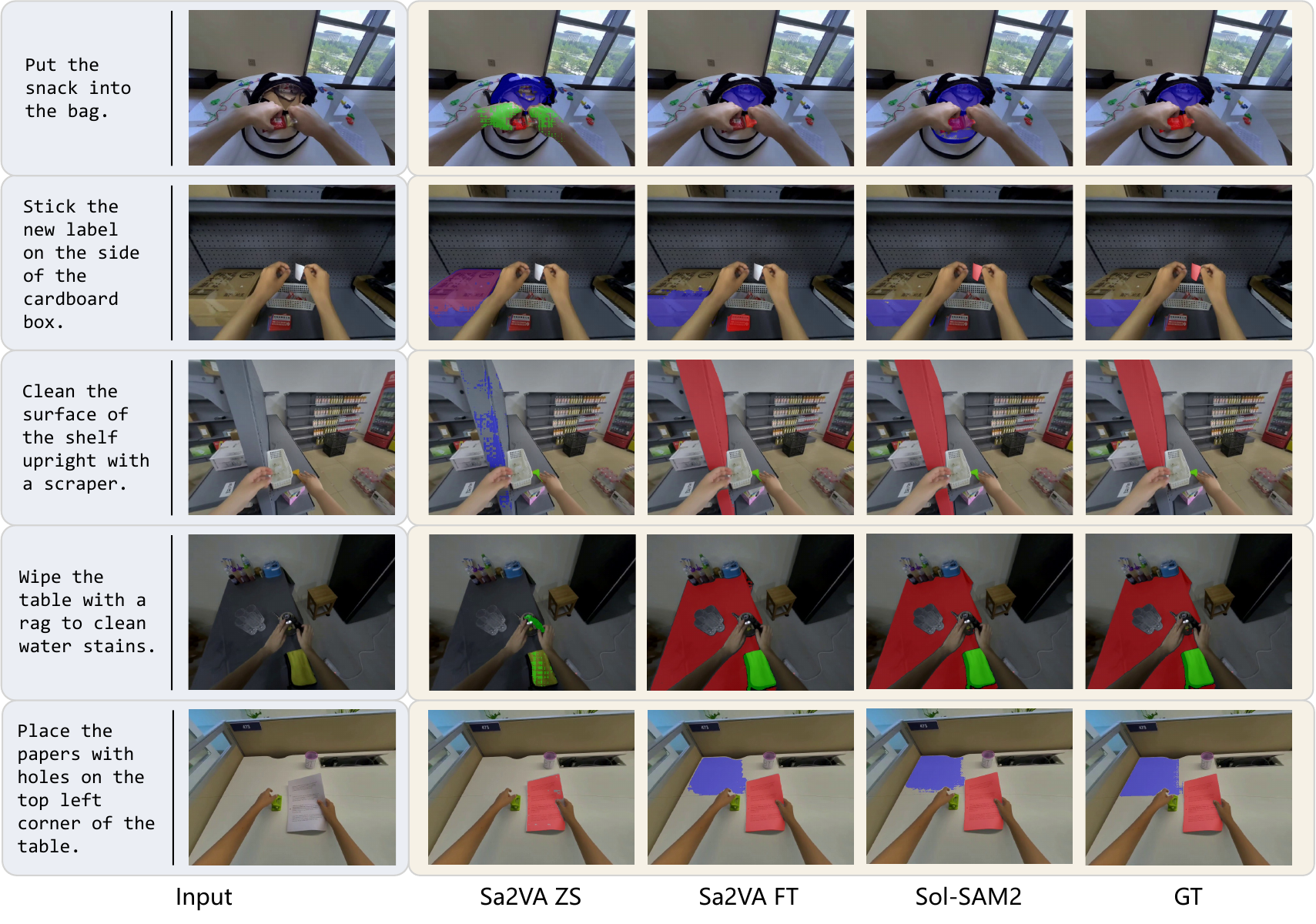}
    \caption{\textbf{Role-structured part grounding before and after EgoAfford training.} Red/green/blue denote O/I/D. Examples show improved role assignment and instrument localization, with remaining instance ambiguity and destination errors.}
    \label{fig:qualitative}
\end{figure*}

\Cref{fig:qualitative} compares Sa2VA before and after training with Sol–SAM2 and GT. Training improves role assignments, including excluding hands from the instrument role and correcting direct-object/destination confusion, and improves instrument localization. Target-instance ambiguity and spatial destination errors remain. Additional comparisons across all seven models are provided in \cref{sec:supp-qualitative}.

\subsection{Benchmark analysis}
\label{sec:benchmark-analysis}
To test action conditioning, we fix images, checkpoints, and original GT, then remove the instruction or replace it with one from another episode in the same domain. A GT-independent replacement mapping is shared across methods. We use Sol–SAM2 for the intervention study to limit API costs, given its similar aggregate scores to Astra–SAM2. On the real test set, removing and mismatching instructions lower cIoU by 33.40--51.86 and 47.41--54.08 points, respectively (\cref{tab:action-conditioning}). These drops show that the supplied action description strongly influences grounding predictions. Synthetic results, paired uncertainty estimates, and output-validity analysis appear in \cref{sec:supp-analysis}.

\begin{table}[htbp]
    \centering
    \caption{\textbf{Action-instruction interventions on the real test set (cIoU, \%).} Images, checkpoints, and original GT are fixed across all 2,215 samples.}
    \label{tab:action-conditioning}
    \small
    \setlength{\tabcolsep}{3pt}
    \begin{tabularx}{\linewidth}{@{}Yrrr@{}}
        \toprule
        Model & Correct & Removed & Mismatched \\
        \midrule
        \multicolumn{4}{l}{\textit{General segmentation models}} \\
        \cmidrule(lr){1-4}
        LENS-3B & 56.43 & 21.04 & 8.67 \\
        UniPixel-7B & 54.65 & 15.26 & 7.24 \\
        Sa2VA-8B & 61.12 & 20.99 & 7.05 \\
        Qwen3-VL-8B-SAMTok & 58.79 & 23.74 & 6.99 \\
        \midrule
        \multicolumn{4}{l}{\textit{Affordance-specialist models}} \\
        \cmidrule(lr){1-4}
        AffordanceVLM-7B & 55.86 & 18.58 & 7.27 \\
        Affordance-R1-7B & 57.24 & 20.72 & 7.63 \\
        AFUN-8B & 58.81 & 6.95 & 7.53 \\
        \midrule
        \multicolumn{4}{l}{\textit{Off-the-shelf pipeline}} \\
        \cmidrule(lr){1-4}
        Sol--SAM2 & 59.00 & 25.60 & 7.66 \\
        \bottomrule
    \end{tabularx}
\end{table}

NE-mIoU averages per-sample IoU over GT-positive cases, with missed predictions scoring zero. Destination regions have the lowest NE-mIoU for all seven trained models (\cref{tab:role-diagnostics}). For the four general models, empty predictions account for 3.20--3.51\% of GT-positive destination cases, highlighting region selection and localization as challenges alongside role binding. The tool pipelines retain stronger direct-object and destination localization, while AFUN has the highest instrument NE-mIoU. All seven trained models have higher presence F1 than the tool pipelines. Further diagnostics appear in \cref{sec:supp-role-diagnostics}.

\begin{table}[htbp]
    \centering
    \caption{\textbf{Role-wise results on the real test set (\%).} NE-mIoU averages IoU over 2,172/423/1,282 GT-positive O/I/D cases; empty predictions score zero. P-F1: macro presence F1 over 2,215 samples. Bold/underline: best/second-best trained model across the seven evaluated frameworks.}
    \label{tab:role-diagnostics}
    \small
    \setlength{\tabcolsep}{3pt}
    \begin{tabularx}{\linewidth}{@{}Yrrrr@{}}
        \toprule
        Model & O & I & D & P-F1 \\
        \midrule
        \multicolumn{5}{l}{\textit{General segmentation models}} \\
        \cmidrule(lr){1-5}
        LENS-3B & \textbf{69.17} & 67.61 & 52.08 & 97.49 \\
        UniPixel-7B & 66.08 & 67.66 & 49.70 & 97.18 \\
        Sa2VA-InternVL3-8B & 68.82 & 68.16 & \textbf{54.03} & \underline{97.54} \\
        Qwen3-VL-8B-SAMTok & 66.39 & 65.11 & 48.74 & 97.37 \\
        \midrule
        \multicolumn{5}{l}{\textit{Affordance-specialist models}} \\
        \cmidrule(lr){1-5}
        AffordanceVLM-7B & 60.00 & 64.59 & 45.51 & 97.08 \\
        Affordance-R1-7B & \underline{68.83} & \underline{69.74} & 48.47 & \textbf{97.61} \\
        AFUN-8B & 67.59 & \textbf{72.66} & \underline{52.18} & 97.09 \\
        \midrule
        \multicolumn{5}{l}{\textit{Off-the-shelf pipelines}} \\
        \cmidrule(lr){1-5}
        Sol--SAM2 & 75.26 & 68.65 & 57.81 & 95.82 \\
        Astra--SAM2 & 76.36 & 70.23 & 59.06 & 95.68 \\
        \bottomrule
    \end{tabularx}
\end{table}

\subsection{Data scale and composition}
\label{sec:experiments-3}
\label{sec:data-composition}
To assess the value of EgoAfford supervision, we vary the amount and source of training data. Doubling real training data from 12,370 to 24,740 samples improves Sa2VA-Qwen3B from 54.78\% to 57.01\% cIoU on the real test set (\cref{tab:data-sources}, top block). Both runs use the same adaptation and 40 epochs.

We then add 13,706 synthetic samples to all 26,250 real samples. Real-test cIoU is maintained (56.35 to 57.04), while synthetic-test cIoU rises from 21.04 to 64.08. The confirmed object vocabulary expands from 686 to 1,322 concepts, and coverage of real-test object concepts rises from 93.2\% to 96.1\%. We therefore combine both sources to broaden concept coverage while preserving performance on the primary real domain.

On the combined 39,956 samples, Sa2VA with the larger InternVL3-8B backbone reaches 61.12\% real-test cIoU and 67.16\% synthetic-test cIoU.

\begin{table}[htbp]
    \centering
    \caption{\textbf{Data scale and composition (cIoU, \%).} $N_R$/$N_S$: real/synthetic training samples. Comparisons within each 3B block share a recipe; the 8B row is the full-data reference with its own selected recipe. Bold marks the better score within each paired block and the 8B reference.}
    \label{tab:data-sources}
    \small
    \setlength{\tabcolsep}{3pt}
    \begin{tabularx}{\linewidth}{@{}Yrrrr@{}}
        \toprule
        Model & $N_R$ & $N_S$ & Real & Synthetic \\
        \midrule
        \multicolumn{5}{l}{\textit{Training-set size}} \\
        \cmidrule(lr){1-5}
        Sa2VA-3B & 12,370 & 0 & 54.78 & 29.97 \\
        Sa2VA-3B & 24,740 & 0 & \textbf{57.01} & \textbf{36.91} \\
        \midrule
        \multicolumn{5}{l}{\textit{Training-set composition}} \\
        \cmidrule(lr){1-5}
        Sa2VA-3B & 26,250 & 0 & 56.35 & 21.04 \\
        Sa2VA-3B & 26,250 & 13,706 & \textbf{57.04} & \textbf{64.08} \\
        \midrule
        \multicolumn{5}{l}{\textit{Full-data 8B reference}} \\
        \cmidrule(lr){1-5}
        Sa2VA-8B & 26,250 & 13,706 & \textbf{61.12} & \textbf{67.16} \\
        \bottomrule
    \end{tabularx}
\end{table}

We also compare the sources at a fixed total of 12,370 training samples (\cref{tab:equal-size-sources}). The training protocol for this comparison and the other data studies is specified in \cref{sec:supp-source-comparison}.

\begin{table}[htbp]
\centering
\caption{\textbf{Equal-size source comparison with Sa2VA-Qwen3B (cIoU, \%).} $N_R+N_S=12,370$ for every row. This study uses its own training recipe.}
\label{tab:equal-size-sources}
\small
\setlength{\tabcolsep}{3pt}
\begin{tabularx}{\linewidth}{@{}Yrrrr@{}}
\toprule
Training & $N_R$ & $N_S$ & Real & Synthetic \\
\midrule
Real & 12,370 & 0 & \underline{54.78} & 29.97 \\
Synthetic & 0 & 12,370 & 36.27 & \underline{62.68} \\
Mixed & 6,185 & 6,185 & \textbf{55.16} & \textbf{65.85} \\
\bottomrule
\end{tabularx}
\end{table}

At a fixed total of 12,370 training samples, mixed training achieves the highest cIoU on both test sets (55.16\% real and 65.85\% synthetic). With half as many real samples as real-only training, it maintains real-domain performance (54.78\% for real-only) while improving synthetic-test cIoU by 35.88 points. These results support the complementary value of real and synthetic supervision.

%% file: sec/7_manipulation.tex
\Needspace{12\baselineskip}
\section{Manipulation studies}
\label{sec:manipulation-studies}
We examine how role-part masks enter two manipulation systems: a learned policy conditioned on region features and an agentic executor given explicit spatial references. The studies connect grounding predictions and role annotations to action execution.

\subsection{Policy conditioning in LIBERO-Goal}
\label{sec:experiments-4}
\begin{figure}[H]
\centering
\includegraphics[width=\linewidth]{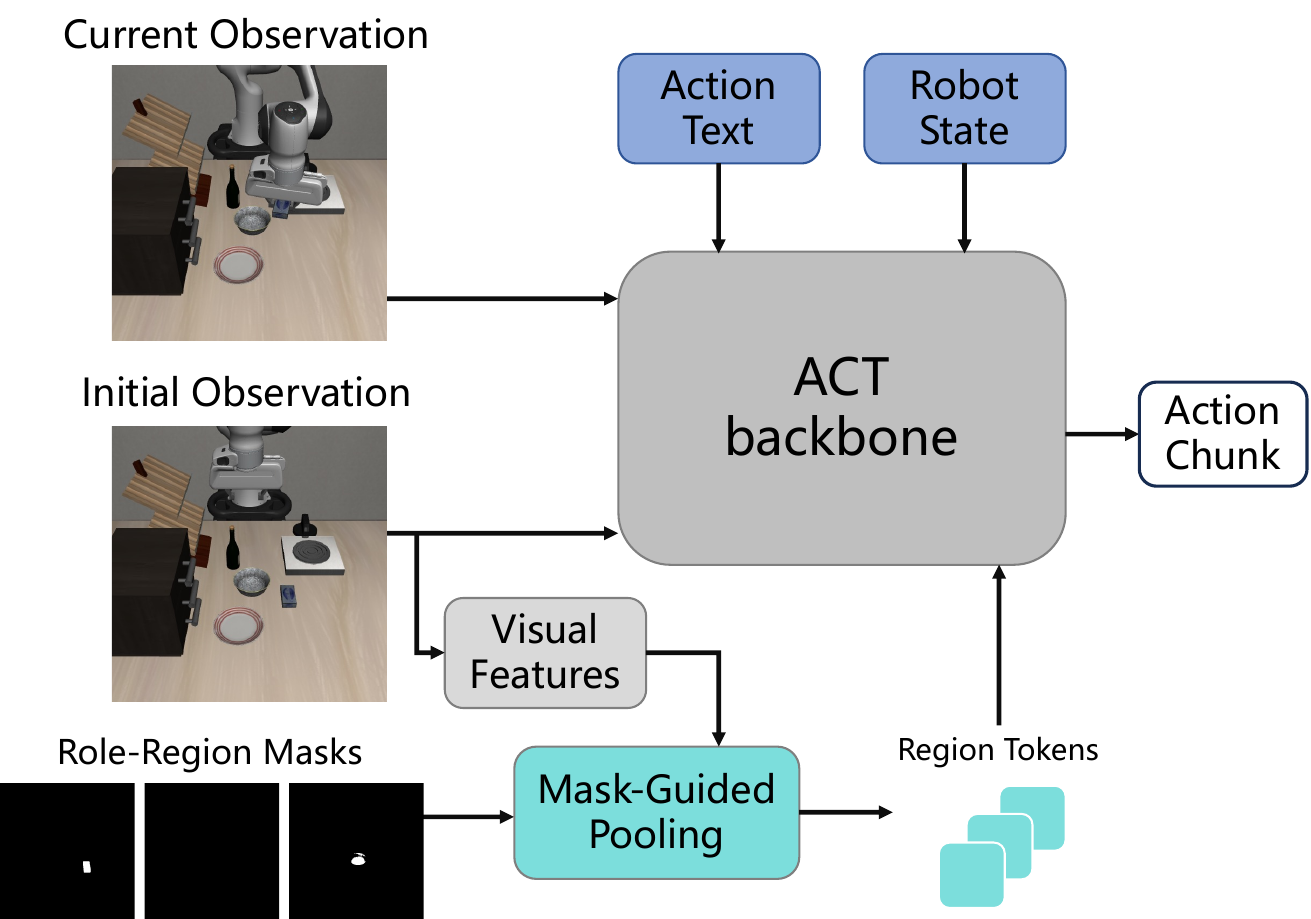}
\caption{\textbf{ACT-derived executor with mask-guided region tokens.} Role masks restrict attention pooling over initial-image features to form region tokens that join the ACT encoder input. Current and initial RGB provide global visual features, while action text and robot state supply context. The policy outputs action chunks; initial RGB and regions remain fixed within a step.}
\label{fig:supp-act-architecture}
\end{figure}

We test whether role-part masks improve a learned policy's execution of a given action. We build an ACT-derived policy that uses masks to select spatial image features and supplies the pooled region tokens to its action encoder (\cref{fig:supp-act-architecture}).

\begin{figure*}[!t]
\centering
\includegraphics[width=\linewidth]{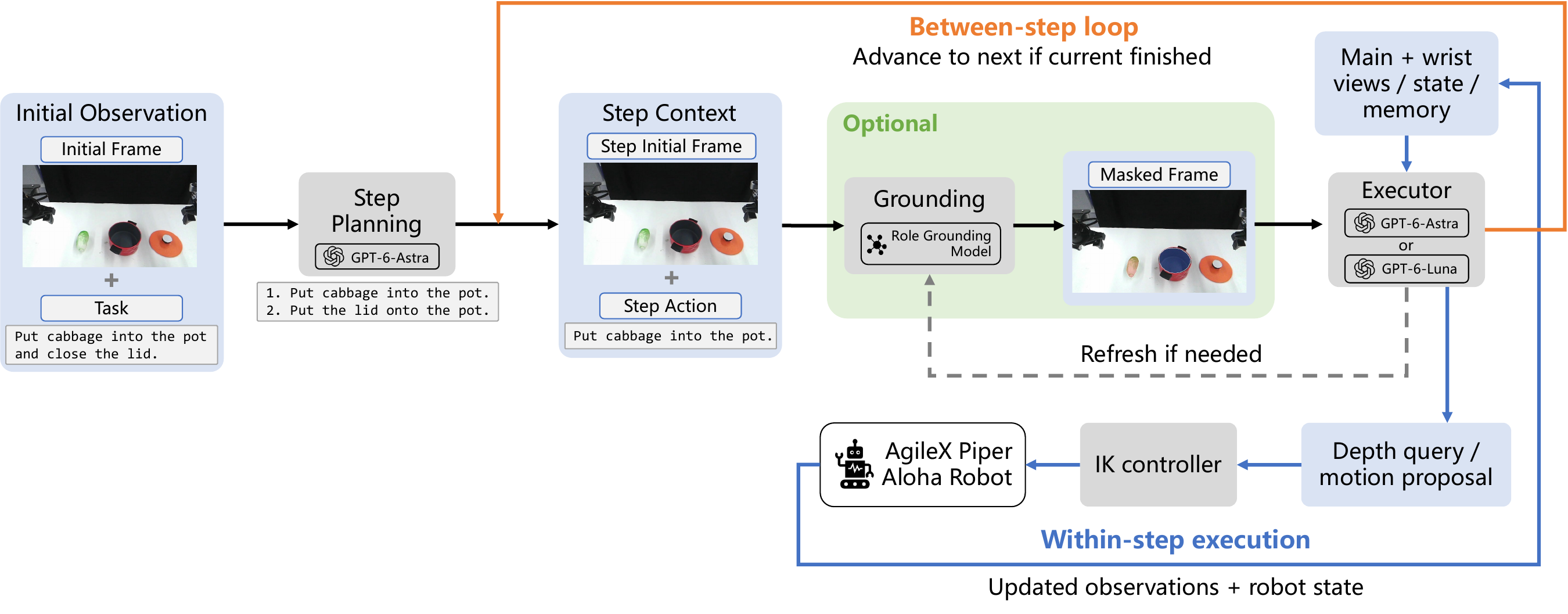}
\caption{\textbf{Role-conditioned agentic manipulation.} Astra plans semantic steps, and the role-grounding module supplies overlays on a step-reference image. The blue loop updates live observations and robot state within a step; the orange loop advances to the next step. The gray dashed path refreshes the role reference when needed. The role-conditioning branch is optional; both conditions share planning, sensing, memory, and control.}
\label{fig:supp-real-robot-system}
\end{figure*}

We start from 500 official LIBERO-Goal~\cite{liu2023libero} expert demonstrations, of which 447 replay successfully. We use 400 replayed demonstrations for training: 35 from one task yield two actions each under our schema, while the other 365 yield one action each, producing 435 training image--action pairs. The remaining 47 demonstrations yield 51 validation image--action pairs under the same annotation procedure. Using these annotations, we train ACT-derived policies~\cite{zhao2023learning} with and without GT role-part masks. We evaluate closed-loop execution on the 51 validation steps, each initialized from its demonstration state, with its action supplied and completion determined by simulator predicates.

Starting from the validation-selected EgoAfford-trained Sa2VA-8B checkpoint, we adapt perception on 435 LIBERO and 435 EgoAfford samples (286 real, 149 synthetic). We update language LoRA, the text-to-mask projection, and the SAM2 mask decoder, selecting the perception checkpoint on the 51 validation samples. Initial-observation masks then condition the fixed GT-trained ACT policy without policy retraining.

\begin{table}[H]
    \centering
    \caption{\textbf{Step-level execution success on 51 validation steps.} The GT and predicted-mask conditions use the same frozen ACT policy. Predicted masks come from domain-adapted Sa2VA-8B. Each step is initialized from its demonstration state.}
    \label{tab:downstream-utility}
    \small
    \setlength{\tabcolsep}{3pt}
    \begin{tabularx}{\linewidth}{@{}Yrr@{}}
        \toprule
        Region input & Successes & Success (\%) \\
        \midrule
        No regions & 41/51 & 80.4 \\
        GT role-part masks & \textbf{45/51} & \textbf{88.2} \\
        Predicted role-part masks & \textbf{45/51} & \textbf{88.2} \\
        \bottomrule
    \end{tabularx}
\end{table}

GT masks yield 45/51 successful steps (88.2\%), versus 41/51 (80.4\%) without regions (\cref{tab:downstream-utility}). Predicted masks also yield 45/51 with the same policy: 44 shared successes, one failure becoming a success, and one success becoming a failure. Learned role-part predictions can therefore condition the fixed executor after perception-domain adaptation. Full input comparisons, metrics, architecture, and per-task results appear in \cref{sec:supp-downstream-act}.

\subsection{Agentic real-robot manipulation}
\label{sec:supp-real-robot}
\label{sec:real-robot}
We test whether explicit role-part masks improve task execution in an agentic manipulation system on a physical robot. Comparing Luna and Astra executors with and without role conditioning lets us examine how this representation benefits executors of different capabilities. \Cref{fig:supp-real-robot-rollouts} shows representative executions of the three tasks. Three demonstration videos accompany the study.

\label{sec:supp-real-robot-system}
The system in \cref{fig:supp-real-robot-system} uses an AgileX dual-arm platform equipped with two 6-DoF Piper manipulators and parallel-jaw grippers, with a main RGB-D view and wrist-camera inputs. GPT-6-astra (Astra) decomposes each task into semantic steps with observable completion criteria. GPT-6-luna (Luna) or Astra then executes the plan using current observations, robot state, and recent memory. Robot-state inputs comprise measured tool-center-point (TCP) poses and gripper openings; recent memory summarizes prior execution and depth queries. Pixel-based depth queries provide calibrated surface coordinates; inverse kinematics and the Piper control stack execute the proposed motions. The action interface adapts RoboDawn's action syntax and parsing components~\cite{guo2026robodawn}.

We compare two input conditions. With role conditioning, the executor receives O/I/D mask overlays on a step-reference image alongside live observations; a fine-tuned Sa2VA-8B model generates these masks online.

Regions can be refreshed as execution progresses. Without role conditioning, the executor receives the language steps and unannotated reference image, with no role descriptions, masks, or grounding output. Both conditions retain the same observation, memory, measurement, and control interfaces.

\paragraph{Tasks and evaluation.}
\label{sec:supp-real-robot-protocol}
We fix ten settings for each task and evaluate four conditions: Luna or Astra execution, each with or without role conditioning. This gives 120 rollouts across the three tasks. Astra generates plans automatically under the same planning procedure across conditions, without condition-specific human guidance. The wording and step decomposition may vary between runs.

The tasks are cabbage placement followed by lid closure, pen placement followed by drawer closure, and sweeping a block into a dustpan. Cabbage and pen tasks award 0.5 per completed action. Sweeping receives 1 when the block ends in the dustpan and 0 otherwise; optional dustpan positioning receives no separate credit. A score of 1 counts as a full-task success. Detailed reference plans, role assignments, and shared execution budgets are given in \cref{sec:appendix-real-robot-protocol}.

\begin{strip}
\captionsetup{hypcap=false}
\centering
\includegraphics[width=\linewidth]{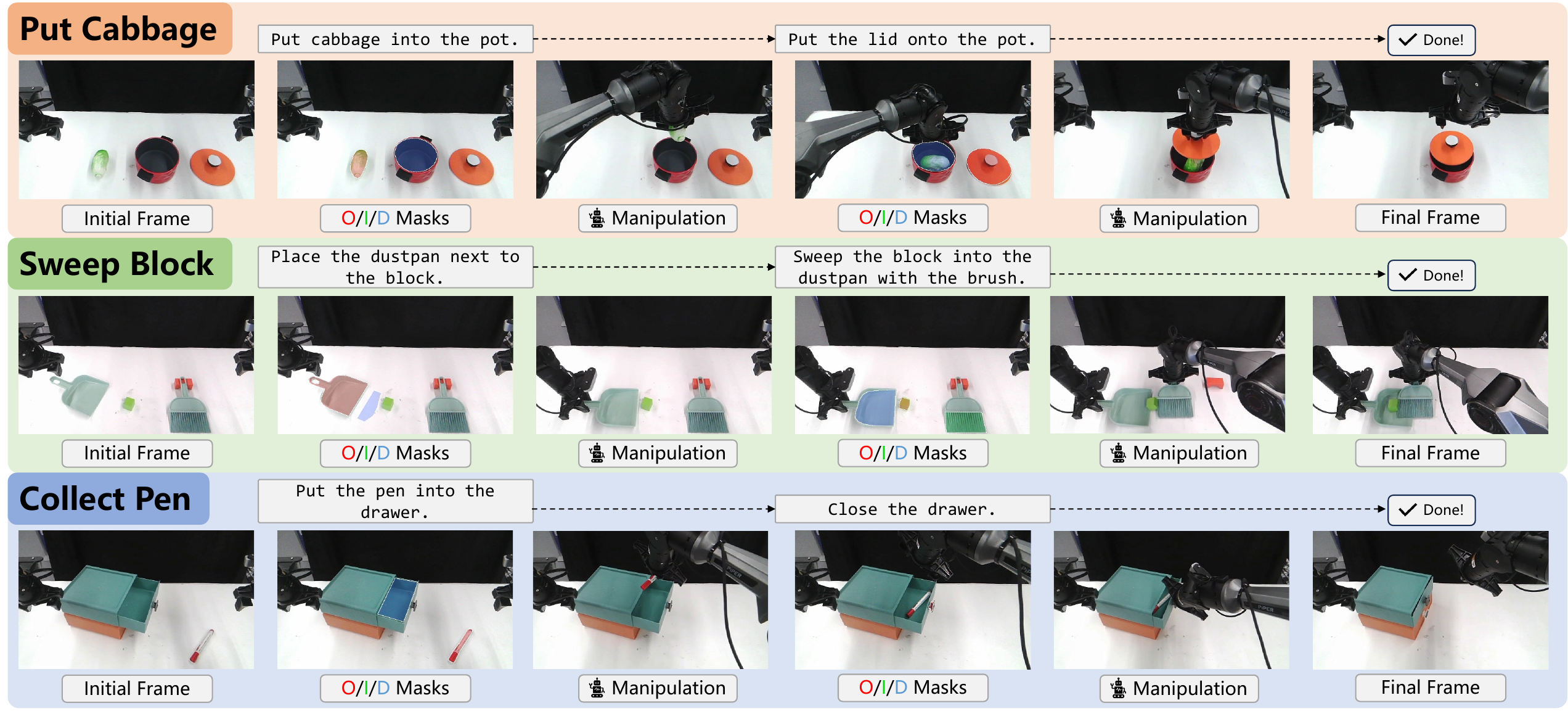}
\captionof{figure}{\textbf{Real-robot manipulation examples.} From top to bottom: cabbage placement and lid closure, dustpan positioning and sweeping, and pen placement and drawer closure. Each row follows the initial observation through step-specific role overlays and robot actions to the final observation. Red, green, and blue denote direct object, instrument, and destination. The overlays change with the action at each step. Demonstration videos accompany the study.}
\label{fig:supp-real-robot-rollouts}
\captionsetup{hypcap=true}
\end{strip}

\paragraph{Results.}
\label{sec:supp-real-robot-results}
Role conditioning raises Luna's mean task-completion score from 35.0\% to 58.3\%, with gains on all three tasks (\cref{tab:supp-real-robot-performance}). Full-task successes increase from 5/30 to 14/30. Astra's mean completion score increases from 78.3\% to 83.3\%, with full-task successes increasing from 21/30 to 22/30.

\begin{table}[H]
\centering
\caption{\textbf{Real-robot task performance.} Mean completion scores are normalized by ten rollouts per task and thirty overall. Cabbage and pen tasks award 0.5 per completed action; sweeping uses binary final-outcome scoring. A full-task success receives 1. All conditions share Astra initial planning.}
\label{tab:supp-real-robot-performance}
\small
\setlength{\tabcolsep}{2.5pt}
\begin{tabularx}{\linewidth}{@{}YYrrrr@{}}
\toprule
Executor & Roles & Cabbage & Pen & Sweep & Mean \\
\midrule
\multicolumn{6}{l}{\textit{Mean task-completion score (\%)}} \\
Luna & Without & 60.0 & 45.0 & 0.0 & 35.0 \\
Luna & With & 85.0 & 70.0 & 20.0 & \textbf{58.3} \\
Astra & Without & 85.0 & 80.0 & 70.0 & 78.3 \\
Astra & With & 80.0 & 90.0 & 80.0 & \textbf{83.3} \\
\midrule
\multicolumn{6}{l}{\textit{Full-task successes}} \\
Luna & Without & 3/10 & 2/10 & 0/10 & 5/30 \\
Luna & With & 7/10 & 5/10 & 2/10 & 14/30 \\
Astra & Without & 7/10 & 7/10 & 7/10 & 21/30 \\
Astra & With & 6/10 & 8/10 & 8/10 & 22/30 \\
\bottomrule
\end{tabularx}
\end{table}

Explicit role-part masks provide actionable spatial references for manipulation, with larger gains for Luna. Together with the learned-mask policy study in \cref{sec:experiments-4}, these results demonstrate the utility of the role-part representation across policy-based and agentic manipulation systems.

Resource-use statistics and the definitions of control rounds and model requests are reported in \cref{sec:appendix-real-robot-resources}.

%% file: sec/8_discussion.tex
\section{Discussion and limitations}
\label{sec:discussion}
EgoAfford brings action-participant roles and functional regions into a shared annotation schema for real and synthetic observations, with a human-reviewed real test set for role-wise, pixel-level evaluation. Its supervision improves grounding across model families, enabling an 8B-backbone model to achieve real-test cIoU comparable to the Astra--SAM2 pipeline. The data studies show benefits from additional real observations and demonstrate how synthetic data extends concept coverage while preserving real-domain performance.
The manipulation studies demonstrate how role-part regions can supply visual references to both a learned policy and an agentic executor.

The benchmark evaluates grounding from a single image and a given action, with coverage shaped by the scenes and viewpoints retained during data construction. Destination regions are harder to localize precisely than the other roles in our evaluation. Expanding scene and interaction coverage and developing models that better capture action-dependent spatial targets are promising directions for more precise localization and cross-domain transfer.

%% file: sec/X_appendix.tex
\onecolumn
\appendix
\begin{center}
{\Large\bfseries Appendix}
\end{center}
\begingroup
\raggedbottom
\crefalias{section}{appendix}
\crefalias{subsection}{appendix}
\crefalias{figure}{appfigure}
\crefalias{table}{apptable}
\crefname{appfigure}{Appendix Fig.}{Appendix Figs.}
\Crefname{appfigure}{Appendix Figure}{Appendix Figures}
\crefname{apptable}{Appendix Tab.}{Appendix Tabs.}
\Crefname{apptable}{Appendix Table}{Appendix Tables}
\crefname{appendix}{Appendix}{Appendices}
\Crefname{appendix}{Appendix}{Appendices}
\input{sec/appendix/1_dataset}
\clearpage
\input{sec/appendix/2_training}

\clearpage
\input{sec/appendix/3_inference}
\clearpage
\input{sec/appendix/4_analysis}
\clearpage
\input{sec/appendix/5_manipulation}
\endgroup

%% file: sec/appendix/1_dataset.tex
\section{Dataset details}
\label{sec:supp-dataset}
\label{sec:supp-annotation}
\label{sec:supp-1}
\Needspace{6\baselineskip}
\subsection{Annotation conventions}
\label{sec:supp-annotation-rules}
\label{sec:supp-real-construction}
\label{sec:supp-synthetic-construction}
\label{sec:supp-data-quality}
\begin{figure}[H]
\centering
\includegraphics[width=\linewidth]{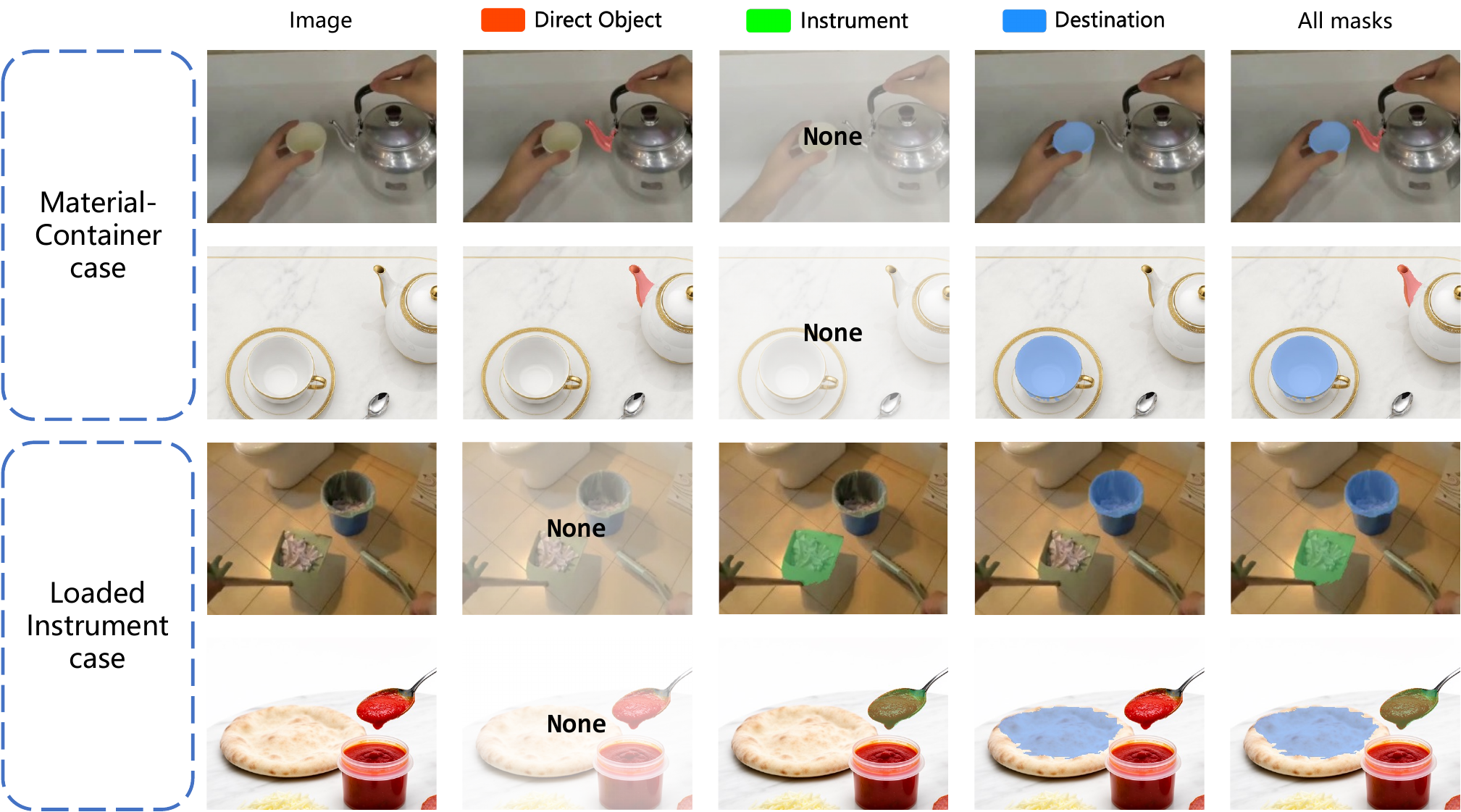}
\caption{\textbf{Material-handling examples.} Top two rows: for an action such as ``pour water into the cup,'' the direct-object mask covers the kettle or teapot's spout rather than the water, and the cup is the destination. Bottom two rows: the dustpan with debris and the spoon with sauce are each annotated as a loaded instrument, including the carried material, with no separate direct-object mask.}
\label{fig:supp-annotation}
\end{figure}

\Cref{fig:supp-annotation} illustrates how the container and loaded-instrument conventions in \cref{sec:task-1} represent manipulated materials.
The same conventions apply to EgoStandard-derived~\cite{lightwheel2026egosuite} and synthetic observations annotated with SAM2~\cite{ravi2024sam2}.
Human review of real-test candidates checks action--image consistency, role assignments, and action-relevant mask boundaries; inconsistent candidates are corrected or rejected.

\clearpage
\subsection{Dataset statistics}
\label{sec:supp-statistics}
\paragraph{Concept mapping.}
\label{sec:supp-concepts}
\label{sec:supp-diversity}
We normalize action and object names using one manually reviewed mapping shared across the two training and two test groups. Synonymous expressions, spelling variants, and inflections are merged while semantic distinctions are retained. For example, \emph{tape gun}, \emph{tape applicator}, and \emph{tape dispenser} map to one concept, while \emph{cup} and \emph{mug} remain distinct. Unresolved and excluded phrases are tracked separately.

Each record contributes at most one action concept. Repeated occurrences of an object concept within that record count once for global object frequencies; role profiles count its named-role occurrences. The mapping is prepared over the vocabulary union for descriptive statistics. \Cref{tab:supp-concept-coverage} reports record coverage by confirmed concepts. Regional descriptions and unresolved names are excluded from this vocabulary count, without removing their samples or masks. Test-candidate review and semantic-name normalization are separate procedures. These percentages measure coverage of the semantic mapping rather than annotation accuracy.
\begin{table}[H]
\centering
\caption{\textbf{Record coverage by the confirmed concept mapping (\%).} Object coverage requires at least one confirmed object concept in a record.}
\label{tab:supp-concept-coverage}
\small
\setlength{\tabcolsep}{4pt}
\begin{tabularx}{\linewidth}{@{}Yrr@{}}
\toprule
Split / source & Action & Object \\
\midrule
Real train & 98.4 & 77.8 \\
Synthetic train & 97.8 & 90.6 \\
Real test & 98.4 & 76.7 \\
Synthetic test & 98.8 & 93.9 \\
\bottomrule
\end{tabularx}
\end{table}

\paragraph{Region scales.}
\label{sec:supp-role-scales}

\begin{figure}[H]
\centering
\includegraphics[width=\linewidth]{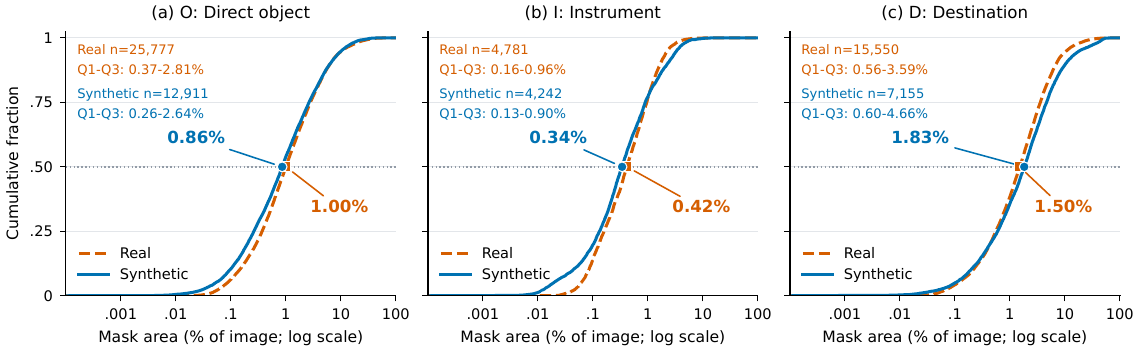}
\caption{\textbf{Training-region area distributions for (a) direct objects, (b) instruments, and (c) destinations.} Curves show empirical CDFs of non-empty GT mask areas, expressed as percentages of image area on a logarithmic axis. Points mark medians; $n$ and Q1--Q3 give the number of non-empty masks and the interquartile range.}
\label{fig:supp-role-scales}
\label{fig:supp-all-role-scales}
\end{figure}

\Cref{fig:supp-role-scales} compares the full distributions of non-empty O/I/D mask areas, normalized by image area, across real and synthetic training. Instruments occupy the smallest regions in both sources, while destinations have the largest median areas.

\begin{figure}[H]
\centering
\includegraphics[width=\linewidth]{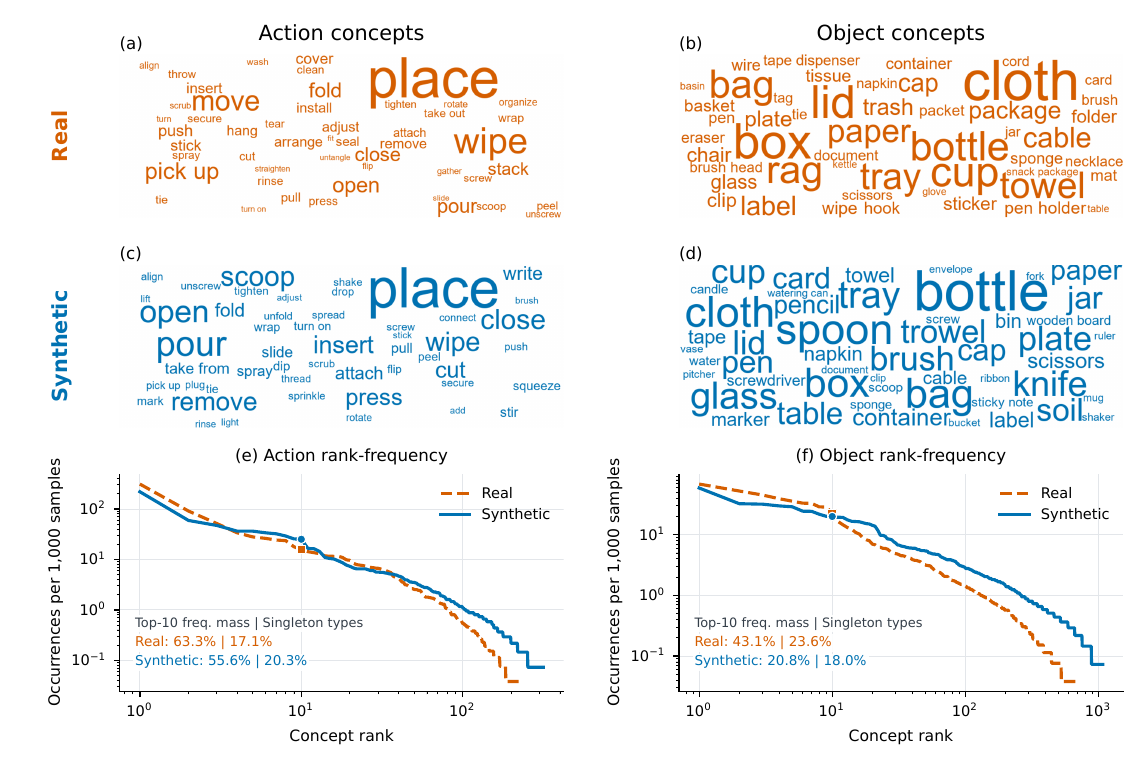}
\caption{\textbf{Semantic profiles of real and synthetic training.} (a--d) Word clouds summarize confirmed action and object concepts, with word size reflecting record counts. (e--f) Rank--frequency curves report records containing each concept per 1,000 source samples, with logarithmic axes. Top-10 frequency mass is the share of concept occurrences contributed by the ten most frequent concepts; singleton types are the fraction of confirmed concepts occurring in exactly one record. Repeated occurrences of an object concept within a record count once.}
\label{fig:supp-concept-profiles}
\end{figure}

\Needspace{0.65\textheight}
\paragraph{Semantic profiles.}
\label{sec:supp-concept-profiles}
\Cref{fig:supp-concept-profiles} combines action and object word clouds with rank--frequency curves for the two training sources. Both contain frequent actions such as \emph{place}, \emph{wipe}, and \emph{pour}, alongside a long tail of less frequent concepts. Real training is more concentrated in its most common concepts: the ten most frequent object concepts account for 43.1\% of confirmed object-concept occurrences, compared with 20.8\% in synthetic training. The broader synthetic vocabulary thus also distributes occurrences across more concepts, complementing the real observations. Named-role profiles are shown in \cref{fig:dataset-coverage}.

\paragraph{Vocabulary coverage.}
\label{sec:supp-vocabulary-coverage}
\Cref{tab:supp-vocabulary-coverage} reports confirmed object-concept coverage and occurrence-weighted coverage of the evaluation sets. The latter weights each concept by the number of test records containing it; unresolved concepts are excluded from both denominators.
\begin{table}[H]
\centering
\caption{\textbf{Confirmed test object-concept coverage (\%), with occurrence-weighted coverage in parentheses.} Mixed training uses the union of the real and synthetic vocabularies.}
\label{tab:supp-vocabulary-coverage}
\small
\setlength{\tabcolsep}{4pt}
\begin{tabularx}{\linewidth}{@{}Yrrr@{}}
\toprule
Evaluation set & Real train & Syn. train & Mixed train \\
\midrule
Real test & 93.2 (98.9) & 79.5 (94.5) & 96.1 (99.4) \\
Synthetic test & 51.2 (73.0) & 84.0 (93.0) & 85.2 (93.5) \\
\bottomrule
\end{tabularx}
\end{table}

%% file: sec/appendix/2_training.tex
\section{Baseline training}
\label{sec:supp-training}
\label{sec:supp-configurations}
\label{sec:supp-2}

\Needspace{6\baselineskip}
\subsection{Adaptation selection}
\label{sec:supp-interfaces}
\label{sec:supp-selection}
\label{sec:supp-adaptation}
\label{sec:supp-4}
\label{sec:supp-selected-recipes}
\label{sec:supp-ablation-lens}
\label{sec:supp-ablation-unipixel}
\label{sec:supp-ablation-sa2va}
\label{sec:supp-ablation-samtok}
\begin{figure}[H]
\centering
\includegraphics[width=0.90\linewidth,height=0.72\textheight,keepaspectratio]{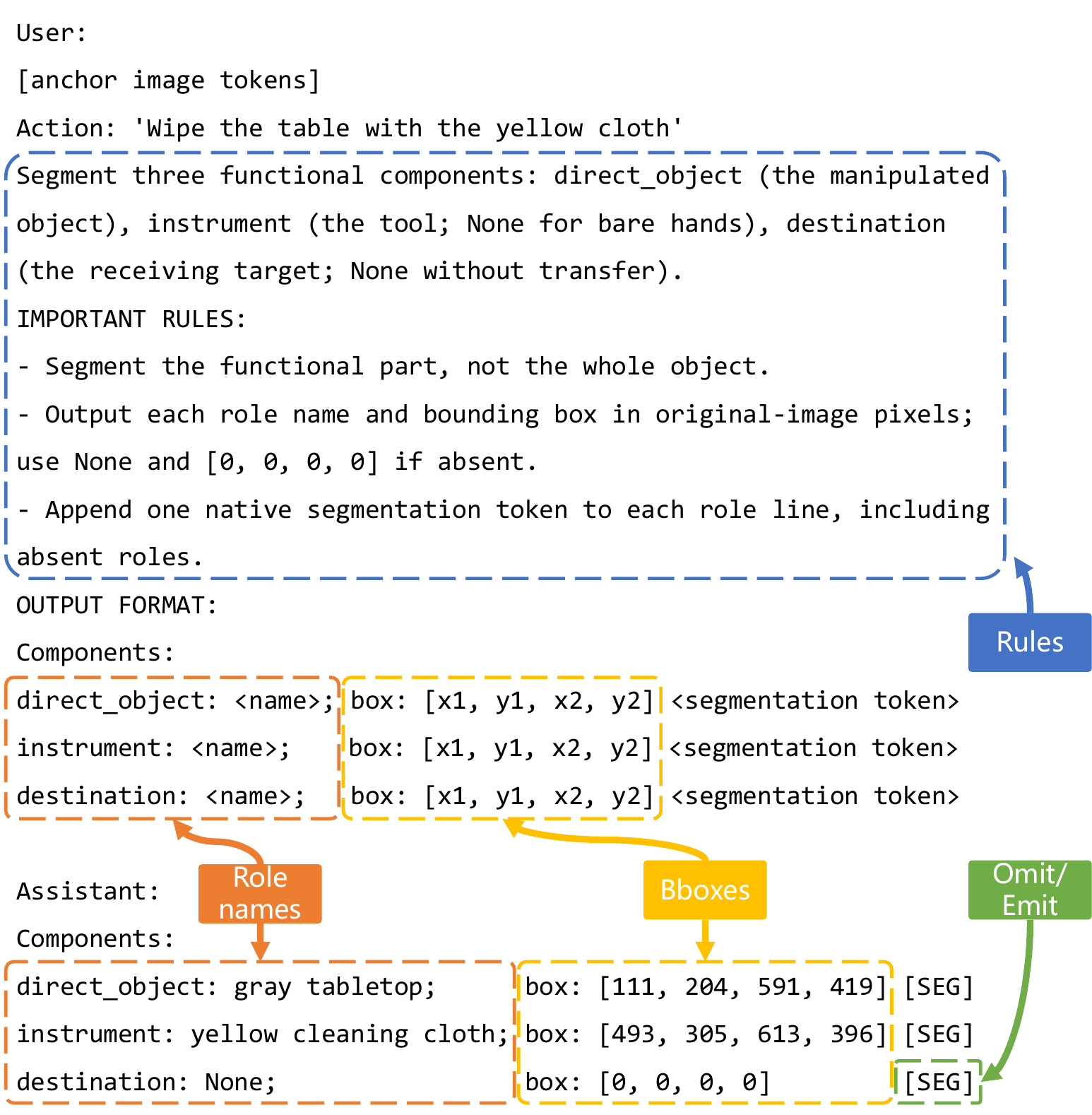}
\caption{\textbf{Training prompt and target with full-rule prompting and emit handling.} Colored outlines mark rule text, participant-name and box targets, and the absent-role \texttt{[SEG]}. Name ablation preserves role identifiers. Minimal prompting removes both rule and output-format instructions. Omit handling retains the \texttt{None} name and zero box while removing the absent role's \texttt{[SEG]} and mask loss.}
\label{fig:supp-training-template}
\end{figure}

\Cref{fig:supp-training-template} illustrates the prompt and supervision fields varied during adaptation selection using a Sa2VA~\cite{yuan2025sa2va} full-rule/emit example. We use the same role-supervision schema for LENS~\cite{zhu2026lens}, UniPixel~\cite{liu2026unipixel}, and Qwen3-VL--SAMTok~\cite{zhou2026samtok}, while retaining model-specific prompts and native mask interfaces (\cref{sec:baselines-2}). We derive the specialist prompts from the same action, O/I/D, and functional-part instructions, converting the targets to role-labeled \texttt{[AFF]} markers and masks for AffordanceVLM, box/point JSON for Affordance-R1, and role-indexed mask queries for AFUN.

\clearpage
The selection subset contains 6,573 real and 3,453 synthetic samples, sampled by source and whole episode. We compare compatible adaptations for each of the seven models using validation cIoU, with 40 subset-training epochs for the general models and 20 for the affordance specialists. The effective batch and optimization settings appear in \cref{tab:supp-training-settings}; \cref{tab:selected-recipes} summarizes the selected interfaces. Test results do not select recipes or checkpoints.
\begin{table}[H]
\centering
\caption{\textbf{Training settings for adaptation selection on 10,026 samples.} Batch denotes the effective global batch. The general models use 40 subset-training epochs and the specialists use 20.}
\label{tab:supp-training-settings}
\small
\setlength{\tabcolsep}{4pt}
\begin{tabularx}{\linewidth}{@{}YrrYY@{}}
\toprule
Model & Batch & Epochs & Learning rate & Warmup / schedule \\
\midrule
\multicolumn{5}{l}{\textit{General segmentation models}} \\
LENS-3B & 512 & 40 & 3e-5 & 100 updates; cosine \\
UniPixel-7B & 32 & 40 & 2e-5; SAM2: 5e-6 & 3\%; cosine \\
Sa2VA-8B & 32 & 40 & 4e-5 & 5\%; cosine \\
Qwen3-VL-8B-SAMTok & 256 & 40 & 2e-5 & 5\%; cosine \\
\midrule
\multicolumn{5}{l}{\textit{Affordance-specialist models}} \\
AffordanceVLM-7B & 32 & 20 & 4e-5 & 5\%; cosine \\
Affordance-R1-7B & 32 & 20 & 2e-5 & 5\%; cosine \\
AFUN-8B & 32 & 20 & 2e-4 & Constant \\
\bottomrule
\end{tabularx}
\par\smallskip
{\figtabnote Microbatch $\times$ GPUs $\times$ accumulation: LENS $16\times8\times4$, UniPixel $4\times8\times1$, Sa2VA $2\times8\times2$, Qwen3-VL--SAMTok $2\times8\times16$. Their language/mask modules are Qwen2.5-VL-3B/SAM2 Hiera-L, Qwen2.5-VL-7B/SAM2.1 Hiera-B+, InternVL3-8B/SAM2 Hiera-L, and Qwen3-VL-8B/SAMTok, respectively.}
\end{table}

\begin{table}[H]
\centering
\caption{\textbf{Selected adaptation recipes after subset training.} Components denotes separately pretrained language and mask components; official denotes jointly trained model weights. The four general models supervise names, boxes and native masks; specialists retain the compatible targets stated below. Validation scores are given in the model-specific tables.}
\label{tab:selected-recipes}
\small
\setlength{\tabcolsep}{4pt}
\begin{tabularx}{\linewidth}{@{}Ylll@{}}
\toprule
Model & Initialization & Prompt & Absent-role output \\
\midrule
LENS-3B & Components & Full-rule & Native \\
UniPixel-7B & Official & Minimal & Omit \\
Sa2VA InternVL3-8B & Components & Full-rule & Emit \\
Qwen3-VL-8B-SAMTok & Official & Full-rule & Native omit \\
\midrule
AffordanceVLM-7B & Official & Part rule & Empty masks \\
Affordance-R1-7B & Official & Part rule & Null geometry \\
AFUN-8B & Official & Part rule & Empty masks \\
\bottomrule
\end{tabularx}
\end{table}

The preferred task-rule prompt and absent-role encoding differ across models: UniPixel selects minimal prompting with omit handling, whereas Sa2VA selects task rules with emit handling. Removing box targets lowers LENS cIoU by 6.31 points under its minimal prompt (\cref{tab:ablation-lens}). Initialization also has model-dependent effects: Sa2VA's two initializations have similar peak cIoU, while Qwen3-VL--SAMTok gains 5.88 points from official initialization under full-rule prompting (\cref{tab:ablation-sa2va,tab:ablation-samtok}).

\begin{table}[H]
\centering
\begin{minipage}[t]{0.48\linewidth}
\centering
\caption{\textbf{LENS-3B adaptation comparisons.} Name and box ablations use the minimal-prompt reference; task-rule removal uses the selected recipe.}
\label{tab:ablation-lens}
\small
\setlength{\tabcolsep}{4pt}
\begin{tabularx}{\linewidth}{@{}Yrrr@{}}
\toprule
Change from reference & cIoU & gIoU & $\Delta$ \\
\midrule
Selected & \textbf{57.15} & \textbf{76.12} & 0.00 \\
w/o task rules & \underline{56.36} & 74.71 & $-$0.80 \\
Minimal prompt: all text targets & \underline{56.36} & 74.71 & 0.00 \\
Minimal prompt: w/o role names & 55.33 & \underline{75.41} & $-$1.03 \\
Minimal prompt: w/o bounding boxes & 50.04 & 72.08 & $-$6.31 \\
\bottomrule
\end{tabularx}
\end{minipage}
\hfill
\begin{minipage}[t]{0.48\linewidth}
\centering
\caption{\textbf{UniPixel-7B adaptation comparisons.} $\Delta$ is relative to the selected recipe.}
\label{tab:ablation-unipixel}
\small
\setlength{\tabcolsep}{4pt}
\begin{tabularx}{\linewidth}{@{}Yrrr@{}}
\toprule
Change from reference & cIoU & gIoU & $\Delta$ \\
\midrule
Selected & \textbf{55.29} & \underline{74.56} & 0.00 \\
+ task rules & 51.69 & 74.13 & $-$3.60 \\
Official $\rightarrow$ components initialization & 54.05 & 74.12 & $-$1.24 \\
Omit $\rightarrow$ emit empty-role tokens & \underline{54.69} & \textbf{74.84} & $-$0.59 \\
\bottomrule
\end{tabularx}
\end{minipage}
\par\medskip
\begin{minipage}[t]{0.48\linewidth}
\centering
\caption{\textbf{Sa2VA-InternVL3-8B adaptation comparisons.} $\Delta$ is relative to the selected recipe.}
\label{tab:ablation-sa2va}
\small
\setlength{\tabcolsep}{4pt}
\begin{tabularx}{\linewidth}{@{}Yrrr@{}}
\toprule
Change from reference & cIoU & gIoU & $\Delta$ \\
\midrule
Selected & \textbf{61.36} & 75.79 & 0.00 \\
w/o task rules & 59.33 & \textbf{76.20} & $-$2.03 \\
Emit $\rightarrow$ omit empty-role tokens & 59.39 & 75.40 & $-$1.97 \\
Components $\rightarrow$ official initialization & \underline{61.11} & \underline{76.12} & $-$0.25 \\
\bottomrule
\end{tabularx}
\end{minipage}
\hfill
\begin{minipage}[t]{0.48\linewidth}
\centering
\caption{\textbf{Qwen3-VL-8B-SAMTok adaptation comparisons.} $\Delta$ is relative to the selected recipe.}
\label{tab:ablation-samtok}
\small
\setlength{\tabcolsep}{4pt}
\begin{tabularx}{\linewidth}{@{}Yrrr@{}}
\toprule
Change from reference & cIoU & gIoU & $\Delta$ \\
\midrule
Selected & \textbf{57.05} & \textbf{75.66} & 0.00 \\
w/o task rules & \underline{56.49} & \underline{75.14} & $-$0.55 \\
Official $\rightarrow$ components initialization & 51.17 & 71.19 & $-$5.88 \\
\bottomrule
\end{tabularx}
\end{minipage}
\par\smallskip
{\figtabnote All scores are validation percentages. gIoU is reported at the cIoU-selected checkpoint; $\Delta$ is the cIoU change from the stated reference, computed before rounding. Bold and underlining mark the best and second-best displayed values per model and metric.}
\end{table}

\Cref{tab:supp-specialist-adaptation} records recipe selection. AffordanceVLM benefits most clearly from removing name/box targets, raising validation cIoU from 37.76\% to 47.82\%. The smaller validation differences for R1 and AFUN determine their selected recipes.

\begin{table}[H]
\centering
\caption{\textbf{Affordance-specialist adaptation comparisons (\%).} Subset runs use 20 epochs; bold marks the validation-selected recipe within each model. Part rule denotes the instruction to segment functional parts.}
\label{tab:supp-specialist-adaptation}
\small
\setlength{\tabcolsep}{4pt}
\textit{Configuration selection on 10,026 mixed samples}\par\smallskip
\begin{tabularx}{\linewidth}{@{}YYr@{}}
\toprule
Model & Subset configuration & Val cIoU \\
\midrule
Affordance-R1 & Names, boxes, points, part rule & 53.89 \\
 & Without part rule & 54.43 \\
 & Without names (selected) & \textbf{54.63} \\
\midrule
AffordanceVLM & Names, boxes, masks, part rule & 37.76 \\
 & Without part rule & 39.98 \\
 & Role labels and masks, part rule (selected) & \textbf{47.82} \\
 & Masks and role labels, without part rule & 46.59 \\
\midrule
AFUN & Independent queries, part rule & 55.88 \\
 & Independent queries, without part rule & 55.18 \\
 & Shared queries, part rule (selected) & \textbf{56.04} \\
\bottomrule
\end{tabularx}

\end{table}

\clearpage
\subsection{Training settings}
\label{sec:supp-full-training}
\label{sec:appendix-specialist-training}
The full-data runs restart from their selected initializations and train on 39,956 samples using the settings in \cref{tab:supp-full-training}, rather than continuing from subset-trained checkpoints.

LENS fully tunes its VLM and connector with the SAM2 image encoder frozen, while UniPixel uses vision/language LoRA with trainable SAM2. Sa2VA and Qwen3-VL--SAMTok use language LoRA with frozen visual encoders; Sa2VA trains the SEG projector and mask decoder with the SAM2 prompt encoder frozen, and Qwen3-VL--SAMTok keeps the SAMTok codec frozen. Token embeddings and LM heads remain trainable in all three LoRA-based models.
Affordance-R1 learns role-wise box/point JSON using cross-entropy and language LoRA, with SAM2 frozen; its selected target omits names. AffordanceVLM learns three role-labeled \texttt{[AFF]} markers and masks, including empty masks, without name/box text. Its language LoRA, embeddings, output head, segmentation projector, and SAM mask decoder are trainable. Both use AdamW. AFUN trains its query embeddings, normalization, and language-to-SAM3 projector with native SAM3 segmentation losses, using Adam.  Its 96 query positions share 32 embeddings across the three roles; Qwen and SAM3 are frozen. Absent roles retain their output slots and receive null geometry or empty-mask supervision as appropriate.

The specialists use effective batch 32 and 40 epochs. Their validation cIoU scores are 56.39\% for AffordanceVLM, 59.58\% for Affordance-R1, and 63.34\% for AFUN. Main-test cIoU is reported in \cref{tab:main-benchmark}; complementary gIoU appears in \cref{tab:supp-main-giou}.

\begin{table}[H]
\centering
\caption{\textbf{Full-data training settings (seed 42).} Batch denotes the effective global batch; LoRA entries give rank/alpha/dropout. LENS completes 39.50 epochs; the other models complete 40 epochs.}
\label{tab:supp-full-training}
\label{tab:supp-full-optimization}
\small
\setlength{\tabcolsep}{4pt}
\begin{tabularx}{\linewidth}{@{}lrrYYY@{}}
\toprule
Model & Batch & Epochs & Learning rate & Warmup / schedule & LoRA \\
\midrule
LENS-3B & 512 & 39.50 & 3e-5 & 100 updates; cosine & \textemdash{} (full fine-tuning) \\
UniPixel-7B & 32 & 40 & 2e-5; SAM2: 5e-6 & 3\%; cosine & 128/256/0.1 \\
Sa2VA-8B & 32 & 40 & 4e-5 & 5\%; cosine & 128/256/0.05 \\
Qwen3-VL-8B-SAMTok & 256 & 40 & 2e-5 & 5\%; cosine & 128/256/0.05 \\
\midrule
AffordanceVLM-7B & 32 & 40 & 4e-5 & 5\%; cosine & 128/256/0.05 \\
Affordance-R1-7B & 32 & 40 & 2e-5 & 5\%; cosine & 128/256/0.05 \\
AFUN-8B & 32 & 40 & 2e-4 & Constant & \textemdash{} (native queries) \\
\bottomrule\end{tabularx}
\par\smallskip
{\figtabnote LENS uses AdamW with weight decay 0. Qwen3-VL--SAMTok uses AdamW with weight decay 0.05 and native cross-entropy supervision. Inference and parsing are described in \cref{sec:supp-inference}.}
\end{table}

\Needspace{6\baselineskip}
\subsection{Data-study protocols}
\label{sec:supp-source-comparison}
The training-set size comparison in \cref{tab:data-sources} starts from official Sa2VA-Qwen3B weights, with task-rule prompts, name/box targets, native mask supervision, and emit handling. Both nested real-data subsets use effective batch 32, seed 42, and 40 epochs. Each run selects its checkpoint by cIoU on the fixed 778-sample real validation set. Fixed epochs entail more optimization updates for the larger subset.

The full-data source comparison uses the adaptation selected for the main benchmark, with a Qwen3B backbone. The full-data 8B reference uses InternVL3-8B and its own selected recipe from \cref{tab:selected-recipes}. All rows evaluate the reviewed 2,215-sample real test set; the two 3B blocks and the 8B reference retain their respective training configurations.

To assess source complementarity at a fixed training-set size, we compare synthetic-only, real-only, and mixed training at a total of 12,370 samples in \cref{tab:equal-size-sources}. All three runs use official Sa2VA-Qwen3B initialization, task-rule prompts, name/box targets, native mask supervision, emit handling, effective batch 32, seed 42, and 40 epochs, with checkpoint selection on the same 778-sample real validation set. All three rows are evaluated on the same reviewed real and synthetic test sets.

%% file: sec/appendix/3_inference.tex
\section{Inference details}
\label{sec:supp-inference}
\label{sec:supp-evaluation}
\label{sec:supp-3}
\label{sec:supp-zero-shot-prompts}

\subsection{Inference protocols}
All benchmark scores use fixed O/I/D correspondence. Recoverable masks remain in evaluation even when ancillary text is malformed; roles without a recoverable prediction receive empty masks. Every sample contributes to the reported scores. The cIoU, gIoU, NE-mIoU, and presence-F1 definitions follow \cref{sec:task-2}.

\label{sec:supp-mask-parsing}

\begin{figure}[H]
\centering
\includegraphics[width=\linewidth]{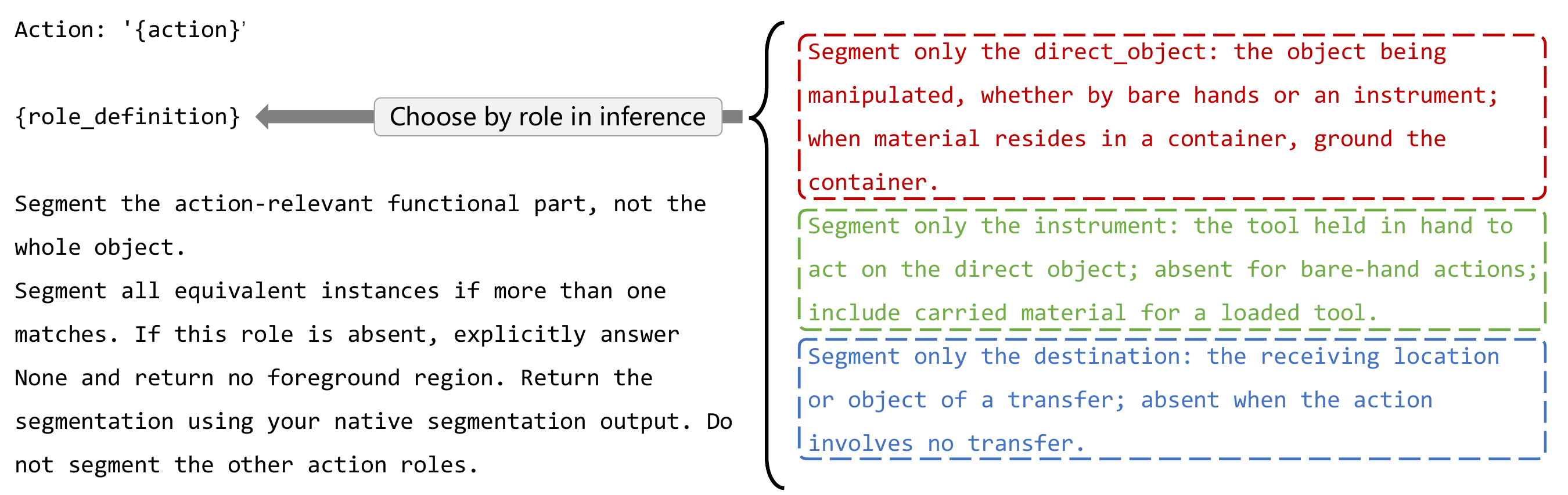}
\caption{\textbf{Role-specific zero-shot prompt template.} Left: the shared template. Right: alternative role definitions. Each independent query inserts exactly one definition into \texttt{\{role\_definition\}}, keeping the action and remaining instructions unchanged. Native prompt wrappers and output formatting remain model-specific.}
\label{fig:supp-zs-prompt}
\end{figure}

To assess whether the four general models can directly produce role-structured outputs from their official checkpoints, we compare joint and role-specific zero-shot queries. Joint queries use the full-rule prompt schema in \cref{fig:supp-training-template}, requesting role names, boxes, and masks together through each model's native interface. LENS has a fixed single-mask decoder, so joint multi-mask results are marked as unsupported. \Cref{tab:joint-zs} compares joint and role-specific inference. For each image--action pair, we instantiate the template in \cref{fig:supp-zs-prompt} three times, replacing \texttt{\{role\_definition\}} with the corresponding O/I/D definition for each independent query before applying the model's native formatting.

\label{sec:supp-specialist-zs}
For zero-shot evaluation, all seven segmentation models use their public weights without EgoAfford training. Each receives the current image and action using the role-specific template in \cref{fig:supp-zs-prompt}, with input formatting adapted to its native interface. Three independent queries produce masks in fixed O/I/D order; no GT names, boxes, or role-presence labels are supplied.

AffordanceVLM from RAGNet~\cite{wu2025ragnet} uses its native SAM decoder and the official single-query mask selection. Affordance-R1~\cite{wang2026affordance} predicts boxes and points for SAM2 Hiera-L; the highest-quality candidate for each prompt is selected and predictions for the same role are merged. AFUN~\cite{wang2026afun} uses its RGB/text segmentation path with SAM3, selecting the highest-scoring detection above confidence 0.5. Its depth and motion branches are not used. Generation limits are 512 tokens for AffordanceVLM and 1,024 for Affordance-R1, compared with 1,536 for the four general models; AFUN has no autoregressive decoding step. AFUN uses the publicly released checkpoint associated with its motion evaluation, which differs from the earlier checkpoint used for its main segmentation table.

For all seven segmentation models, masks are selected without GT assistance and evaluated against the original-resolution EgoAfford annotations.

\paragraph{VLM--SAM2.}
\label{sec:supp-vlm-sam}
\begin{figure}[H]
\centering
\includegraphics[width=\linewidth]{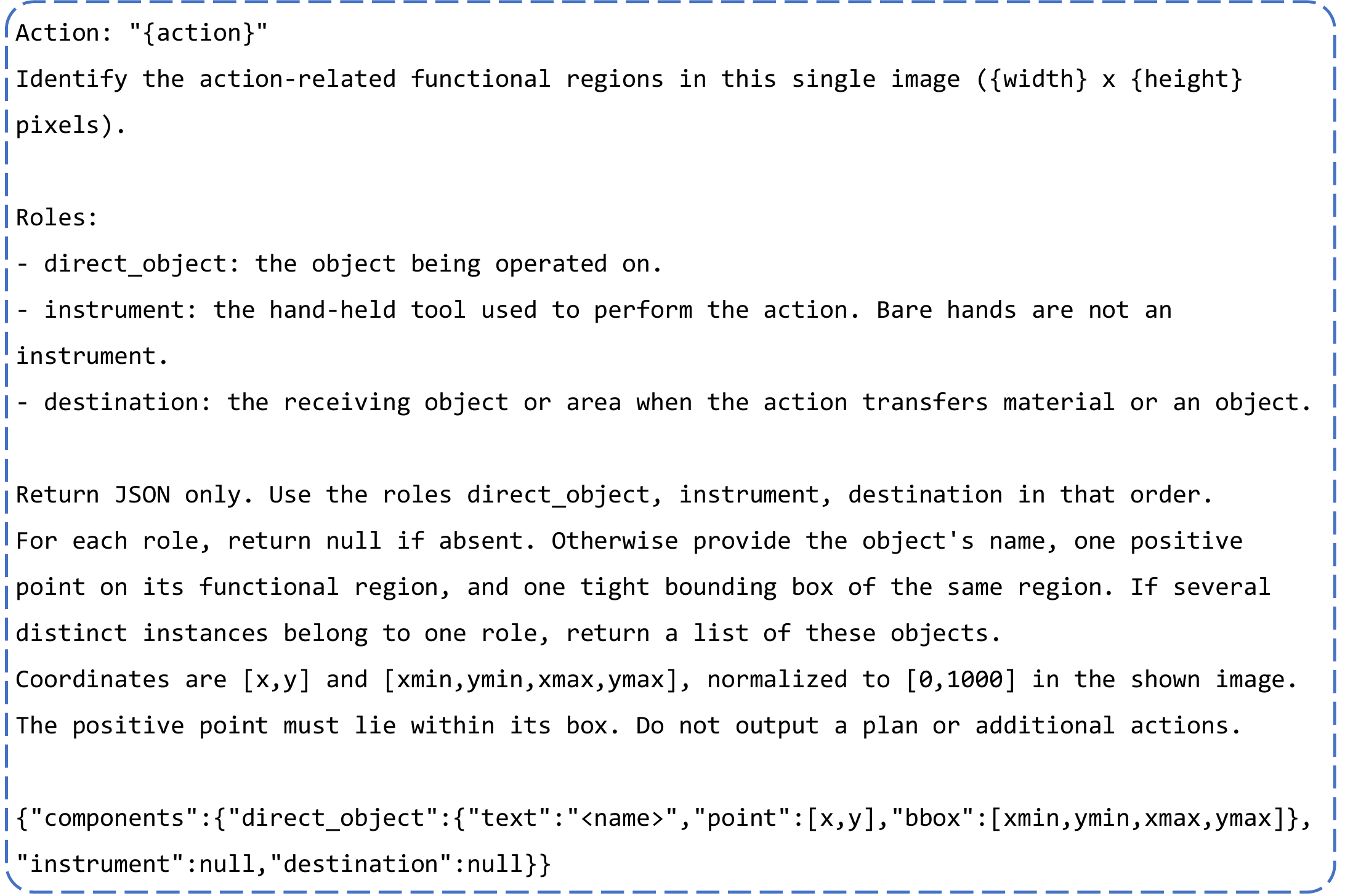}
\caption{\textbf{VLM--SAM2 prompt template.} Given the image and action, the VLM returns role-specific names, positive points, and boxes, with \texttt{null} for absent roles and lists for multiple instances. The geometric cues prompt SAM2 to produce masks.}
\label{fig:supp-vlm-sam-prompt}
\end{figure}
\Cref{fig:supp-vlm-sam-prompt} shows the VLM query template. It requests role-wise object names, positive points, and bounding boxes in JSON, using \texttt{null} for absent roles. Coordinates are normalized to $[0,1000]$ and converted to image coordinates for SAM2.

When Sol or Astra produces an incomplete or malformed response, we retain parseable content and repair its format before extracting masks, allowing at most one retry if parsing or geometry validation still fails, without using GT or evaluation scores to select responses.

\clearpage
\subsection{Zero-shot output compatibility}
\label{sec:supp-joint-zs}
Complete-response compatibility requires a mask or explicit absence declaration for every role, with unambiguous role assignment and valid native formatting. Explicitly absent roles may omit masks, and multiple instance masks may be merged within a role. For separate queries, an image is compatible when all three responses meet this criterion. For both joint and separate inference, we also report the fraction of individual role outputs that are recoverable as a role-bound native mask or explicit absence declaration (\cref{tab:zs-protocol-compatibility}). Both measures concern output usability; an incorrect absence declaration can still be compatible.
\begin{table}[H]
\centering
\caption{\textbf{Zero-shot cIoU (\%) with joint and role-specific queries.} Joint-output failures can obscure zero-shot grounding ability, motivating this comparison with separate role queries. Each role-specific result uses three independent queries per image. All samples are scored with fixed role alignment.}
\label{tab:joint-zs}
\small
\setlength{\tabcolsep}{4pt}
\begin{tabularx}{\linewidth}{@{}Yrrrr@{}}
\toprule
Model & Real: joint & Real: separate & Syn.: joint & Syn.: separate \\
\midrule
LENS-3B & -- & 15.28 & -- & 19.33 \\
UniPixel-7B & 0.00 & 21.02 & 0.00 & 31.97 \\
Sa2VA-InternVL3-8B & 0.00 & 23.05 & 0.00 & 34.53 \\
Qwen3-VL-8B-SAMTok & 4.23 & 27.48 & 1.19 & 41.63 \\
\bottomrule
\end{tabularx}
\par\smallskip
\begin{minipage}{\linewidth}\figtabnote
--: joint multi-mask output is unsupported by LENS's native single-mask decoder.
\end{minipage}
\end{table}
\begin{table}[H]
\centering
\caption{\textbf{Zero-shot protocol compatibility and recoverable role outputs (\%).} This diagnostic tests whether unusable joint role outputs account for the low scores and motivate separate-query evaluation. Compatibility requires a complete O/I/D response per image (2,215 real; 488 synthetic). Recoverable counts usable native masks or explicit absence declarations over role slots (three per image).}
\label{tab:zs-protocol-compatibility}
\label{tab:zs-protocol-mismatch} 
\small
\setlength{\tabcolsep}{3pt}
\begin{tabularx}{\linewidth}{@{}Yrrrr@{}}
\toprule
 & \multicolumn{2}{c}{Real} & \multicolumn{2}{c}{Synthetic} \\
\cmidrule(lr){2-3}\cmidrule(lr){4-5}
Model & Compatibility & Recoverable & Compatibility & Recoverable \\
\midrule
\multicolumn{5}{l}{\textit{Joint zero-shot queries}} \\
\cmidrule(lr){1-5}
LENS-3B & -- & -- & -- & -- \\
UniPixel-7B & 0.00 & 0.00 & 0.00 & 0.00 \\
Sa2VA-InternVL3-8B & 0.00 & 0.00 & 0.00 & 0.00 \\
Qwen3-VL-8B-SAMTok & 7.31 & 57.64 & 3.28 & 40.92 \\
\midrule
\multicolumn{5}{l}{\textit{Separate role-specific zero-shot queries}} \\
\cmidrule(lr){1-5}
LENS-3B & 100.00 & 100.00 & 100.00 & 100.00 \\
UniPixel-7B & 100.00 & 100.00 & 100.00 & 100.00 \\
Sa2VA-InternVL3-8B & 100.00 & 100.00 & 100.00 & 100.00 \\
Qwen3-VL-8B-SAMTok & 56.39 & 82.65 & 77.66 & 92.28 \\
\bottomrule
\end{tabularx}
\par\smallskip
\begin{minipage}{\linewidth}\figtabnote
--: unsupported native interface. Roles follow generated labels (joint) or queried roles (separate), without GT-assisted matching. Explicit absence counts as usable independently of semantic correctness. Missing role outputs remain empty masks; all images are scored.
\end{minipage}
\end{table}

UniPixel and Sa2VA return single-target masks under joint prompting without the requested role bindings. Qwen3-VL--SAMTok can generate role names and boxes but often omits their native mask codes. In separate queries, its recoverable role-output rates are 82.65\% and 92.28\% on the real and synthetic test sets, exceeding the corresponding complete-response compatibility rates of 56.39\% and 77.66\%. Partial responses therefore contribute their available role masks, while missing outputs receive empty masks. The joint runs have no detected generation truncation or native marker/code corruption.

The prompts differ between protocols, so this comparison does not isolate inference count alone.

\begin{table}[H]
\centering
\caption{\textbf{Native output diagnostics for specialist zero-shot role queries (\%).} Syntax validity is measured over 6,645 real and 1,464 synthetic role queries; it is not the image-level complete-response compatibility metric above. AFUN has no autoregressive response syntax, so its no-detection rate is reported separately.}
\label{tab:specialist-zs-format}
\small
\begin{tabularx}{\linewidth}{@{}Yrrrr@{}}
\toprule
 & \multicolumn{2}{c}{Syntax validity} & \multicolumn{2}{c}{No detection} \\
\cmidrule(lr){2-3}\cmidrule(lr){4-5}
Model & Real & Synthetic & Real & Synthetic \\
\midrule
AffordanceVLM-7B & 100.00 & 100.00 & -- & -- \\
Affordance-R1-7B & 97.73 & 94.06 & -- & -- \\
AFUN-8B & -- & -- & 1.13 & 0.14 \\
\bottomrule
\end{tabularx}
\par\smallskip
{\figtabnote R1 has 151/6,645 and 87/1,464 syntax failures; AFUN has 75/6,645 and 2/1,464 no-detection queries. All predictions remain scored, including recoverable regions from malformed responses. No model has detected truncation.}
\end{table}

Specialist format diagnostics in \cref{tab:specialist-zs-format} separate native syntax from detection and semantic correctness. AffordanceVLM's readable outputs still yield low zero-shot grounding scores, while AFUN's no-detection outputs receive empty masks.

\Needspace{8\baselineskip}
\subsection{Additional role-wise results}
\begin{table}[H]
\centering
\caption{\textbf{Role-wise zero-shot results (\%).} O/I/D columns report NE-mIoU over GT-positive cases (real: 2,172/423/1,282; synthetic: 440/191/339), including zero scores for missed predictions. P-F1 is macro presence F1 over all samples and all three roles. Trained-model role scores are reported in \cref{tab:role-diagnostics}.}
\label{tab:supp-zs-roles}
\small
\setlength{\tabcolsep}{3pt}
\begin{tabularx}{\linewidth}{@{}Yrrrrrrrr@{}}
\toprule
 & \multicolumn{4}{c}{Real} & \multicolumn{4}{c}{Synthetic} \\
\cmidrule(lr){2-5}\cmidrule(lr){6-9}
Model & O & I & D & P-F1 & O & I & D & P-F1 \\
\midrule
\multicolumn{9}{l}{\textit{General segmentation models (zero-shot)}} \\
LENS-3B & 37.86 & 37.32 & 17.03 & 68.14 & 32.92 & 37.17 & 24.24 & 77.72 \\
UniPixel-7B & 41.06 & 45.24 & 30.95 & 68.14 & 40.75 & 44.59 & 49.85 & 77.69 \\
Sa2VA-InternVL3-8B & 36.90 & 43.98 & 29.45 & 66.48 & 42.46 & 49.21 & 47.66 & 79.27 \\
Qwen3-VL-8B-SAMTok & 51.36 & 47.14 & 30.27 & 71.99 & 49.02 & 48.97 & 58.79 & 80.43 \\
\midrule
\multicolumn{9}{l}{\textit{Affordance-specialist models (zero-shot)}} \\
AffordanceVLM-7B & 17.12 & 22.00 & 9.32 & 67.99 & 16.36 & 13.62 & 15.11 & 77.69 \\
Affordance-R1-7B + SAM2 & 48.97 & 55.45 & 28.00 & 68.31 & 43.49 & 67.08 & 37.86 & 80.29 \\
AFUN-8B + SAM3 & 35.09 & 50.73 & 21.67 & 67.82 & 27.89 & 8.76 & 31.75 & 77.59 \\
\bottomrule
\end{tabularx}
\end{table}

\Cref{tab:main-benchmark} reports direct-transfer cIoU on both test sets; complementary gIoU appears in \cref{tab:supp-main-giou}. Qwen3-VL--SAMTok has the highest cIoU among these seven zero-shot models. Role-wise results reveal complementary strengths: Affordance-R1 reaches 55.45\% instrument NE-mIoU on the real test set, exceeding the four general zero-shot models; AFUN reaches 50.73\% (\cref{tab:supp-zs-roles}). Role absence remains challenging: on GT-absent real-test role queries, AffordanceVLM, Affordance-R1, and AFUN correctly return empty predictions in 0.11\%, 4.37\%, and 1.12\% of cases, respectively. These results distinguish localization on present roles from suppressing predictions for absent roles.

We extend the action-conditioning and role-localization analyses in \cref{sec:benchmark-analysis} to these adapted models, testing how they use action descriptions and distinguish the three roles. Results are reported in \cref{tab:supp-action-interventions,tab:supp-role-candidates,tab:supp-action-intervals}, with shared-case qualitative comparisons in \cref{fig:supp-qualitative}. AffordanceVLM uses native single-sample inference throughout selection and final evaluation.

%% file: sec/appendix/4_analysis.tex
\section{Benchmark analysis}
\label{sec:supp-analysis}
\label{sec:supp-role-diagnostics}
\Needspace{6\baselineskip}
\subsection{Complementary benchmark scores}
\label{sec:supp-main-metrics}
To complement pixel-accumulated cIoU with sample-level evaluation that includes absent roles, \cref{tab:supp-main-giou} supplements the main benchmark in \cref{tab:main-benchmark} with gIoU for the same checkpoints and test samples. Checkpoint selection remains based on validation cIoU. Sa2VA has the highest real-test gIoU among the trained baselines.
\begin{table}[H]
\centering
\caption{\textbf{Benchmark results (gIoU, \%).} Before: official role-specific zero-shot queries; after: selected task-adapted training. Bold/underline mark best/second-best within each before/after column; tool references are listed separately.}
\label{tab:supp-main-giou}
\small
\setlength{\tabcolsep}{5pt}
\begin{tabularx}{\linewidth}{@{}Yrrrr@{}}
\toprule
& \multicolumn{2}{c}{Real} & \multicolumn{2}{c}{Synthetic} \\
\cmidrule(lr){2-3}\cmidrule(lr){4-5}
Model & Before & After & Before & After \\
\midrule
LENS-3B & 18.04 & \underline{77.79} & 20.43 & 79.06 \\
UniPixel-7B & 22.27 & 76.32 & 29.61 & 79.43 \\
Sa2VA-InternVL3-8B & \underline{27.72} & \textbf{78.33} & 34.59 & 80.78 \\
Qwen3-VL-8B-SAMTok & \textbf{47.39} & 76.20 & \textbf{42.38} & \underline{80.85} \\
\midrule
\multicolumn{5}{l}{\textit{Affordance-specialist models}} \\
AffordanceVLM-7B & 8.84 & 73.23 & 10.19 & 77.07 \\
Affordance-R1-7B & 26.76 & 77.27 & \underline{36.39} & \textbf{83.80} \\
AFUN-8B & 19.35 & \underline{77.79} & 16.88 & 80.79 \\
\midrule
\multicolumn{5}{l}{\textit{Off-the-shelf pipelines (no EgoAfford training)}} \\
\cmidrule(lr){1-5}
GPT-6.1-sol + SAM2 & \multicolumn{2}{c}{79.47} & \multicolumn{2}{c}{79.42} \\
GPT-6-astra + SAM2 & \multicolumn{2}{c}{80.07} & \multicolumn{2}{c}{79.54} \\
\bottomrule
\end{tabularx}
\end{table}

To examine recognition of role presence separately from mask localization, \cref{tab:supp-role-presence} supplements \cref{tab:role-diagnostics} with the role-wise F1 scores underlying its macro average.
\begin{table}[H]
\centering
\caption{\textbf{Role-presence F1 on all 2,215 samples in the real test set (\%).} Macro-F1 is the unweighted mean of O/I/D F1.}
\label{tab:supp-role-presence}
\small
\setlength{\tabcolsep}{4pt}
\begin{tabularx}{\linewidth}{@{}Yrrrr@{}}
\toprule
Model & O F1 & I F1 & D F1 & Macro-F1 \\
\midrule
LENS-3B & 99.65 & 95.64 & 97.18 & 97.49 \\
UniPixel-7B & 99.52 & 94.89 & 97.13 & 97.18 \\
Sa2VA-InternVL3-8B & 99.68 & 95.44 & 97.49 & 97.54 \\
Qwen3-VL-8B-SAMTok & 99.58 & 95.27 & 97.25 & 97.37 \\
\midrule
AffordanceVLM-7B & 99.59 & 94.38 & 97.26 & 97.08 \\
Affordance-R1-7B & 99.70 & 95.72 & 97.41 & 97.61 \\
AFUN-8B & 99.44 & 96.09 & 95.74 & 97.09 \\
\bottomrule
\end{tabularx}
\end{table}

\clearpage
\subsection{Action conditioning}
\label{sec:supp-action-interventions}
To test whether action dependence extends across data sources and prior affordance training, \cref{tab:supp-action-interventions} extends the real-test comparison in \cref{tab:action-conditioning} to both test domains for the seven adapted models and Sol--SAM2. The correct-instruction condition reuses the main benchmark predictions. Instruction substitution uses a seed-42 one-to-one permutation within each domain, excluding donors from the same episode and identical normalized instructions. Models receive neither the original instruction nor the donor identity. The shared fixed-image and original-GT setup follows \cref{tab:action-conditioning}. All predictions, including malformed and all-empty outputs, are scored using the rules in \cref{sec:supp-evaluation}. UniPixel's synthetic no-instruction condition contains 47 format-invalid responses and 109 all-empty predictions, all retained in evaluation.
\begin{table}[H]
\centering
\caption{\textbf{Action-instruction interventions (cIoU, \%).} Images, checkpoints, and original GT remain fixed. The same instruction-replacement mapping is used across models.}
\label{tab:supp-action-interventions}
\small
\setlength{\tabcolsep}{4pt}
\begin{tabularx}{\linewidth}{@{}Yrrrrrr@{}}
\toprule
 & \multicolumn{3}{c}{Real} & \multicolumn{3}{c}{Synthetic} \\
\cmidrule(lr){2-4}\cmidrule(lr){5-7}
Model & Correct & Removed & Mismatched & Correct & Removed & Mismatched \\
\midrule
LENS-3B & 56.43 & 21.04 & 8.67 & 60.72 & 22.06 & 10.39 \\
UniPixel-7B & 54.65 & 15.26 & 7.24 & 59.86 & 4.67 & 11.37 \\
Sa2VA-8B & 61.12 & 20.99 & 7.05 & 67.16 & 16.20 & 11.75 \\
Qwen3-VL-8B-SAMTok & 58.79 & 23.74 & 6.99 & 64.27 & 23.37 & 10.60 \\
Sol--SAM2 & 59.00 & 25.60 & 7.66 & 71.22 & 35.23 & 8.05 \\
\midrule
\multicolumn{7}{l}{\textit{Adapted affordance-specialist models}} \\
AffordanceVLM-7B & 55.86 & 18.58 & 7.27 & 59.04 & 11.83 & 10.57 \\
Affordance-R1-7B & 57.24 & 20.72 & 7.63 & 66.25 & 15.79 & 10.63 \\
AFUN-8B & 58.81 & 6.95 & 7.53 & 62.41 & 9.05 & 12.98 \\
\bottomrule
\end{tabularx}
\end{table}

The specialists also depend strongly on the action: removing it lowers real-test cIoU by 36.52--51.86 points. R1 and AffordanceVLM retain complete parsing without truncation on this domain, while AFUN uses fixed queries. Removing the action reduces destination NE-mIoU from 45.51--52.18\% to 0.00--6.38\%, highlighting the instruction's role in identifying the receiving target.

\Needspace{0.75\textheight}
\subsection{Role localization}
\label{sec:supp-role-mismatch}
To identify predictions that localize another participant more closely than their assigned role, we screen cross-role mask overlaps. For a predicted role $i$ and GT role $j$, let
\begin{equation}
C_{ij}=\operatorname{IoU}(\hat M_i,M_j).
\end{equation}
A potential mismatch pair satisfies $i\ne j$, both masks are non-empty, $C_{ij}\ge0.30$, and $C_{ij}-C_{ii}\ge0.15$. A sample enters the candidate set if at least one pair qualifies and is counted once. The denominator is the full test set, including absent roles and empty or invalid predictions. Pairs whose two GT roles overlap above 0.25 are flagged for review and retained in the count.

\begin{figure}[H]
\centering
\includegraphics[width=\linewidth]{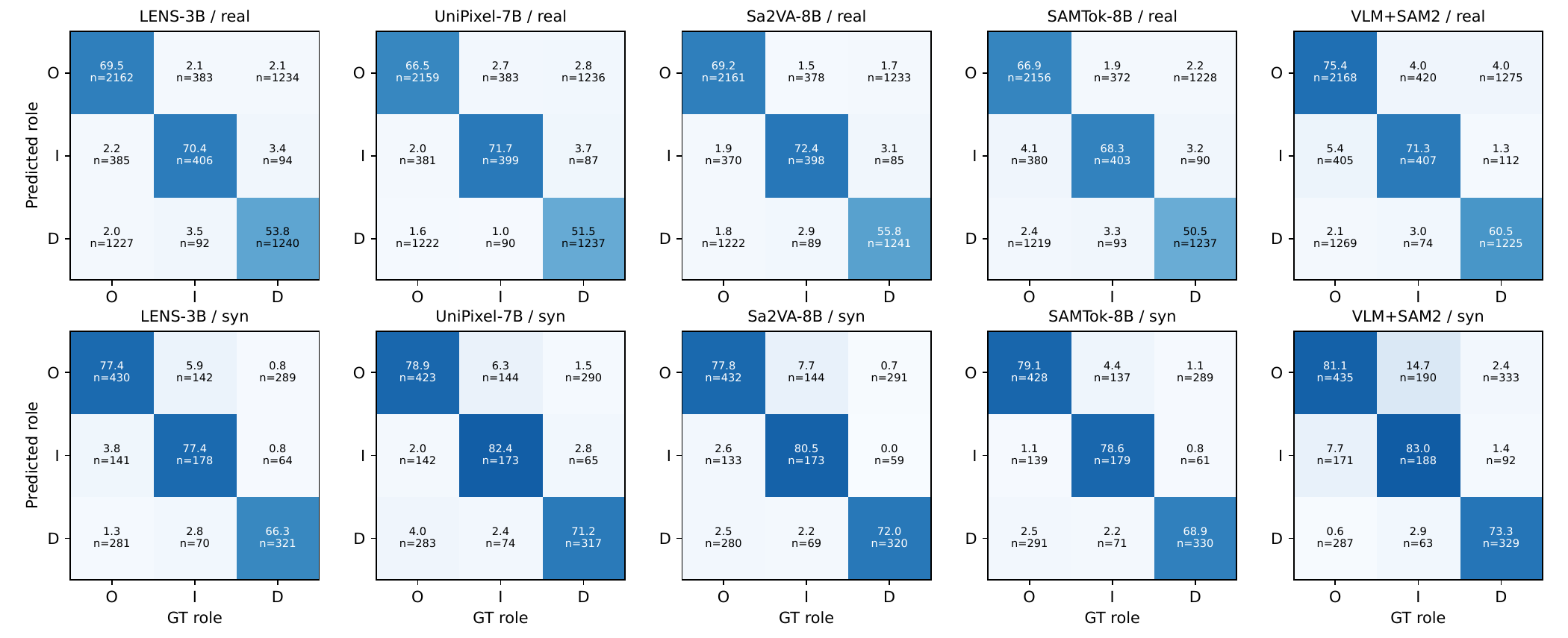}
\caption{\textbf{Cross-role mask overlap across both test domains.} Top/bottom rows show real/synthetic test results for all five methods. Within each matrix, rows are predicted roles and columns are GT roles in O/I/D order. Cells report mean IoU (\%) and support count $n$ over pairs of non-empty predicted and GT masks. ``SAMTok-8B'' denotes Qwen3-VL-8B-SAMTok; ``VLM+SAM2'' denotes Sol--SAM2.}
\label{fig:supp-cross-role-overlap}
\end{figure}

\Cref{tab:supp-role-candidates} reports automatically screened candidates. The thresholds identify cases for inspection; they are not an individually verified role-error rate. The cross-role matrices in \cref{fig:supp-cross-role-overlap} cover the four general baselines and Sol--SAM2 on both test domains. Each cell averages $C_{ij}$ over all test samples with both the predicted row role and GT column role non-empty, and reports its support count; the matrices are not restricted to mismatch candidates. They measure overlap rather than normalized class confusion. Their diagonal differs from NE-mIoU because missed predictions are excluded by this conditioning.
\begin{table}[H]
\centering
\caption{\textbf{Role-mismatch candidate samples, with percentages in parentheses.} Each sample is counted once per model under the stated screening criteria.}
\label{tab:supp-role-candidates}
\label{tab:role-mismatch-candidates}
\small
\setlength{\tabcolsep}{4pt}
\begin{tabularx}{\linewidth}{@{}Yrr@{}}
\toprule
Model & Real test & Synthetic test \\
\midrule
LENS-3B & 59/2215 (2.66) & 19/488 (3.89) \\
UniPixel-7B & 71/2215 (3.21) & 28/488 (5.74) \\
Sa2VA-InternVL3-8B & 55/2215 (2.48) & 22/488 (4.51) \\
Qwen3-VL-8B-SAMTok & 84/2215 (3.79) & 17/488 (3.48) \\
GPT-6.1-sol + SAM2 & 112/2215 (5.06) & 56/488 (11.48) \\
\midrule
\multicolumn{3}{l}{\textit{Adapted affordance-specialist models}} \\
AffordanceVLM-7B & 120/2215 (5.42) & 26/488 (5.33) \\
Affordance-R1-7B & 70/2215 (3.16) & 16/488 (3.28) \\
AFUN-8B & 83/2215 (3.75) & 27/488 (5.53) \\
\bottomrule
\end{tabularx}
\end{table}

\Needspace{0.35\textheight}
\paragraph{Destination localization.}
\label{sec:supp-destination-localization}
To characterize destination localization beyond the per-sample scores in \cref{tab:role-diagnostics}, \cref{tab:supp-destination-localization} reports role-wise micro-IoU, computed by accumulating intersection and union within each role over the full test set, alongside destination-omission counts. These accumulated scores differ from the per-sample NE-mIoU reported in the main text.
\begin{table}[H]
\centering
\caption{\textbf{Role-wise micro-IoU and destination omissions on the real test set (\%).} Micro-IoU is accumulated per role and differs from NE-mIoU. Omission counts use the 1,282 GT-positive destination cases.}
\label{tab:supp-destination-localization}
\small
\setlength{\tabcolsep}{4pt}
\begin{tabularx}{\linewidth}{@{}Yrrrr@{}}
\toprule
Model & O & I & D & D omissions \\
\midrule
LENS-3B & 65.23 & 66.33 & 44.07 & 42/1,282 (3.28\%) \\
UniPixel-7B & 63.21 & 72.19 & 42.84 & 45/1,282 (3.51\%) \\
Sa2VA-InternVL3-8B & 69.45 & 74.83 & 48.96 & 41/1,282 (3.20\%) \\
Qwen3-VL-8B-SAMTok & 68.18 & 68.14 & 45.85 & 45/1,282 (3.51\%) \\
GPT-6.1-sol + SAM2 & 74.16 & 58.86 & 43.11 & 57/1,282 (4.45\%) \\
\bottomrule
\end{tabularx}
\end{table}

Destination micro-IoU is lower than that of direct objects and instruments for every model. Empty predictions account for 3.20--4.45\% of GT-positive destination cases, indicating that destination errors also involve selecting and delineating the target region.

\Needspace{0.50\textheight}
\subsection{Uncertainty across test groups}
\label{sec:supp-action-uncertainty}
We examine whether the large cIoU drops from removing or mismatching action instructions in \cref{tab:action-conditioning,tab:supp-action-interventions} remain stable under changes in test-set composition. Specifically, we quantify how much the correct-minus-intervened cIoU gap varies when different combinations of test episodes or scenes are sampled.

Using the existing predictions, we group real samples into 1,712 source episodes and synthetic samples into 100 generation scenes. Each of 3,000 paired bootstrap resamples draws groups with replacement, using the same sampled groups for the correct and intervened conditions. We recompute cIoU from their accumulated intersections and unions and measure the difference between conditions. The 2.5th and 97.5th percentiles of these differences give the 95\% intervals in \cref{tab:supp-action-intervals}.

For example, removing the instruction lowers Sa2VA's real-test cIoU by 40.13 points, with a 95\% interval of [36.99, 43.26]. All thirty-two intervals, including the specialist results, lie above zero. These results support the main-text finding that removing or mismatching the action causes substantial performance drops: the effect remains pronounced across resampled compositions of the test episodes and scenes.
\begin{table}[H]
\centering
\caption{\textbf{Paired cIoU drops in percentage points, with 95\% group-bootstrap intervals.} Removed and mismatched each compare against the correct instruction.}
\label{tab:supp-action-intervals}
\small
\setlength{\tabcolsep}{4pt}
\begin{tabularx}{\linewidth}{@{}Yrrrr@{}}
\toprule
Model & Real: removed & Real: mismatched & Syn.: removed & Syn.: mismatched \\
\midrule
LENS-3B & 35.39 [32.39, 38.48] & 47.76 [44.75, 50.79] & 38.66 [31.76, 45.34] & 50.33 [44.31, 56.83] \\
UniPixel-7B & 39.39 [36.29, 42.45] & 47.41 [44.50, 50.28] & 55.18 [49.12, 61.87] & 48.49 [42.50, 55.19] \\
Sa2VA-InternVL3-8B & 40.13 [36.99, 43.26] & 54.08 [51.28, 56.84] & 50.96 [44.27, 57.13] & 55.41 [47.41, 63.09] \\
Qwen3-VL-8B-SAMTok & 35.05 [31.69, 38.37] & 51.79 [48.93, 54.55] & 40.90 [34.74, 47.16] & 53.67 [45.50, 61.89] \\
GPT-6.1-sol + SAM2 & 33.40 [29.97, 36.88] & 51.34 [48.30, 54.42] & 35.99 [29.49, 42.39] & 63.17 [56.46, 69.45] \\
\midrule
AffordanceVLM-7B & 37.28 [34.53, 40.07] & 48.59 [45.96, 51.13] & 47.21 [39.65, 55.08] & 48.47 [41.80, 55.74] \\
Affordance-R1-7B & 36.52 [33.33, 39.80] & 49.60 [46.87, 52.44] & 50.46 [43.20, 57.49] & 55.61 [47.82, 63.86] \\
AFUN-8B & 51.86 [48.93, 54.79] & 51.28 [48.48, 54.06] & 53.36 [45.47, 61.43] & 49.43 [41.28, 57.67] \\
\bottomrule
\end{tabularx}
\end{table}

\clearpage
\subsection{Qualitative examples}
\label{sec:supp-qualitative}
To compare adaptation across model families on shared cases, \cref{fig:supp-qualitative} extends \cref{fig:qualitative} to the remaining general baselines, all three specialists, and Astra--SAM2. Training improves role separation across several model families. In the snack-placement example, LENS and UniPixel more clearly distinguish the snack from the receiving bag, while the Sa2VA comparison in \cref{fig:qualitative} shows suppression of the hand-as-instrument prediction. In the shelf-cleaning example, several zero-shot models assign the upright to the destination role or select the hand-held region as the direct object; adaptation instead identifies the upright as the operated surface. The wiping example similarly shows cleaner separation of the tabletop and rag after training in both general and specialist models.

\begin{figure}[H]
\centering
\includegraphics[width=0.98\linewidth]{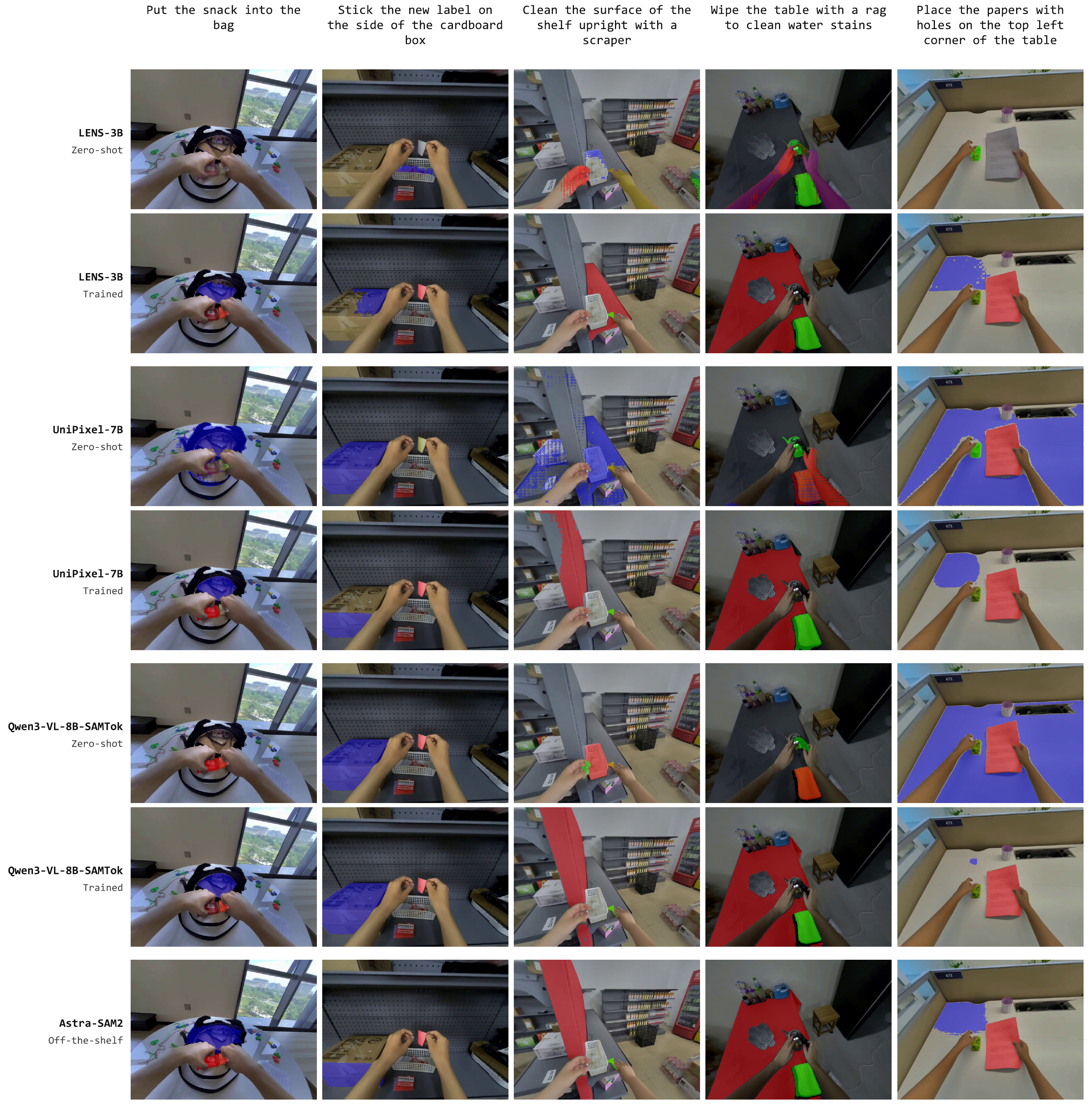}
\caption{\textbf{Additional qualitative comparisons across model families.} Columns show the five cases in the same order as the rows of \cref{fig:qualitative}. Paired rows compare zero-shot and trained predictions from LENS, UniPixel, and Qwen3-VL-8B-SAMTok; the last row shows Astra--SAM2. RGB inputs, GT, Sa2VA and Sol--SAM2 predictions, and the mask color convention are provided in \cref{fig:qualitative}. Continued on the next page.}
\label{fig:supp-qualitative}
\end{figure}

\clearpage
The expanded comparison also reveals differences in instance selection and spatial extent. In the label example, trained Sa2VA and AFUN select another label instance rather than the new label held in the hand. For paper placement, UniPixel and Qwen3-VL--SAMTok initially predict broad destination masks over the table. Training narrows UniPixel's prediction toward the requested upper-left region, whereas Qwen3-VL--SAMTok retains very little destination coverage. Sol--SAM2 and Astra--SAM2 produce well-separated surface and tool masks in the cleaning cases, but differ in the receiving region selected for the bag-placement action. Adapted R1 and AFUN also better separate the snack and bag in this case, while AffordanceVLM places the destination on the tabletop. These examples show how task adaptation changes role assignments and region selection, alongside the remaining ambiguity in target instances and destination boundaries.

\begin{figure}[H]
\ContinuedFloat
\centering
\includegraphics[width=\linewidth]{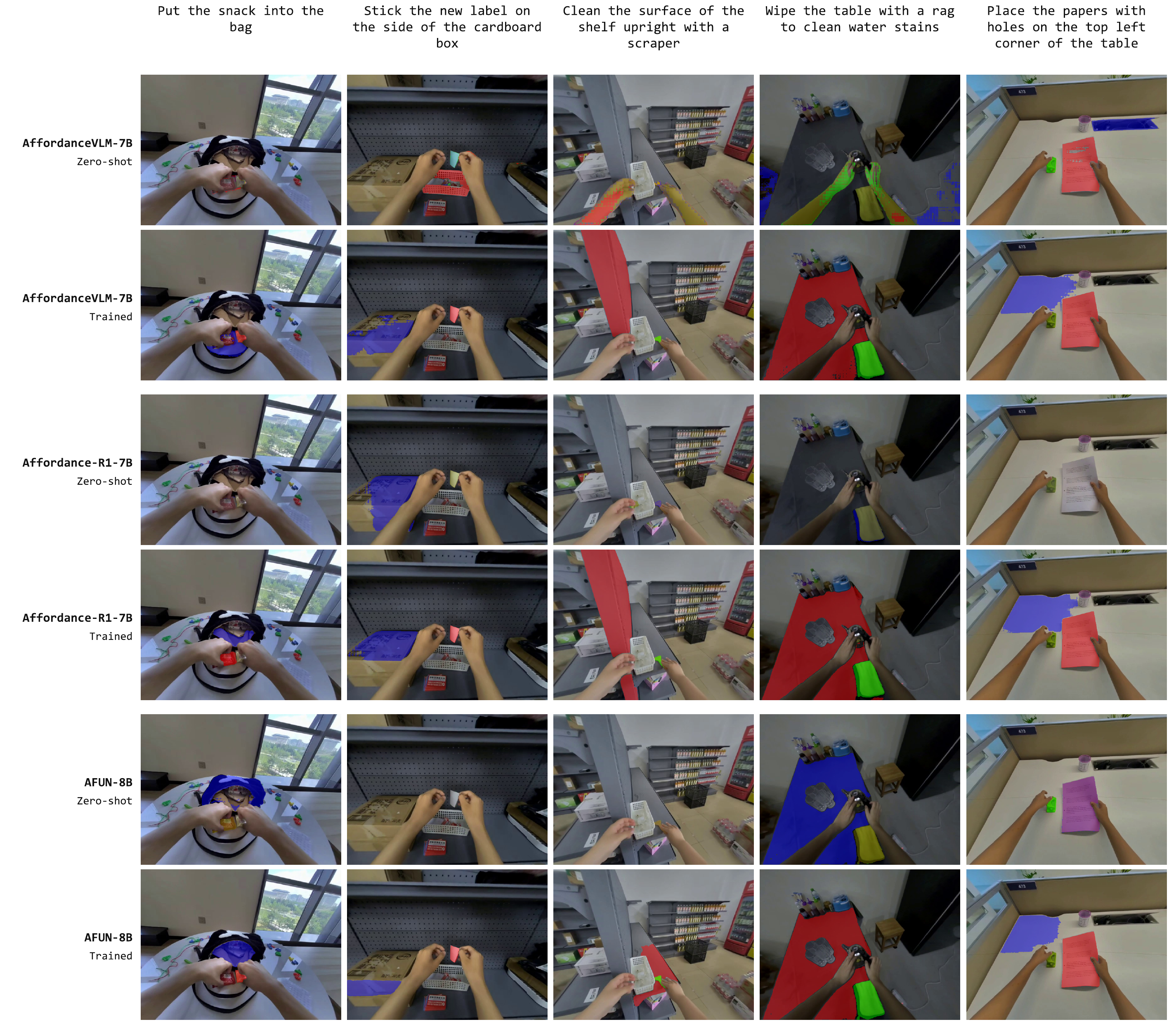}
\caption{\textbf{Additional qualitative comparisons across model families (continued).} Paired rows show zero-shot and adapted AffordanceVLM, Affordance-R1, and AFUN predictions. Columns use the same five cases as the preceding panel; RGB, GT, and mask colors follow \cref{fig:qualitative}. Specialist training and adaptation are detailed in \cref{sec:supp-training}.}
\end{figure}

%% file: sec/appendix/5_manipulation.tex
\section{Manipulation details}
\label{sec:supp-downstream-act}

\Needspace{6\baselineskip}
\subsection{Data and inputs}
\label{sec:supp-act-data}
The LIBERO-Goal~\cite{liu2023libero} demonstration split is specified in \cref{sec:experiments-4}. Training and validation demonstrations are disjoint; failed expert replays are excluded during preparation rather than counted as policy-execution failures.

Each step has an action description and role-part masks on its initial agent-view image. Action windows at each timestep yield 49,172 training and 5,651 validation windows. Each window predicts up to 16 future actions without crossing a step boundary, with padding excluded from the action loss. Action and proprioceptive normalization statistics use training data. The 51 evaluation steps contain 51 direct objects and 37 destinations, with no instruments.

\begin{figure}[H]
\centering
\includegraphics[width=0.88\linewidth]{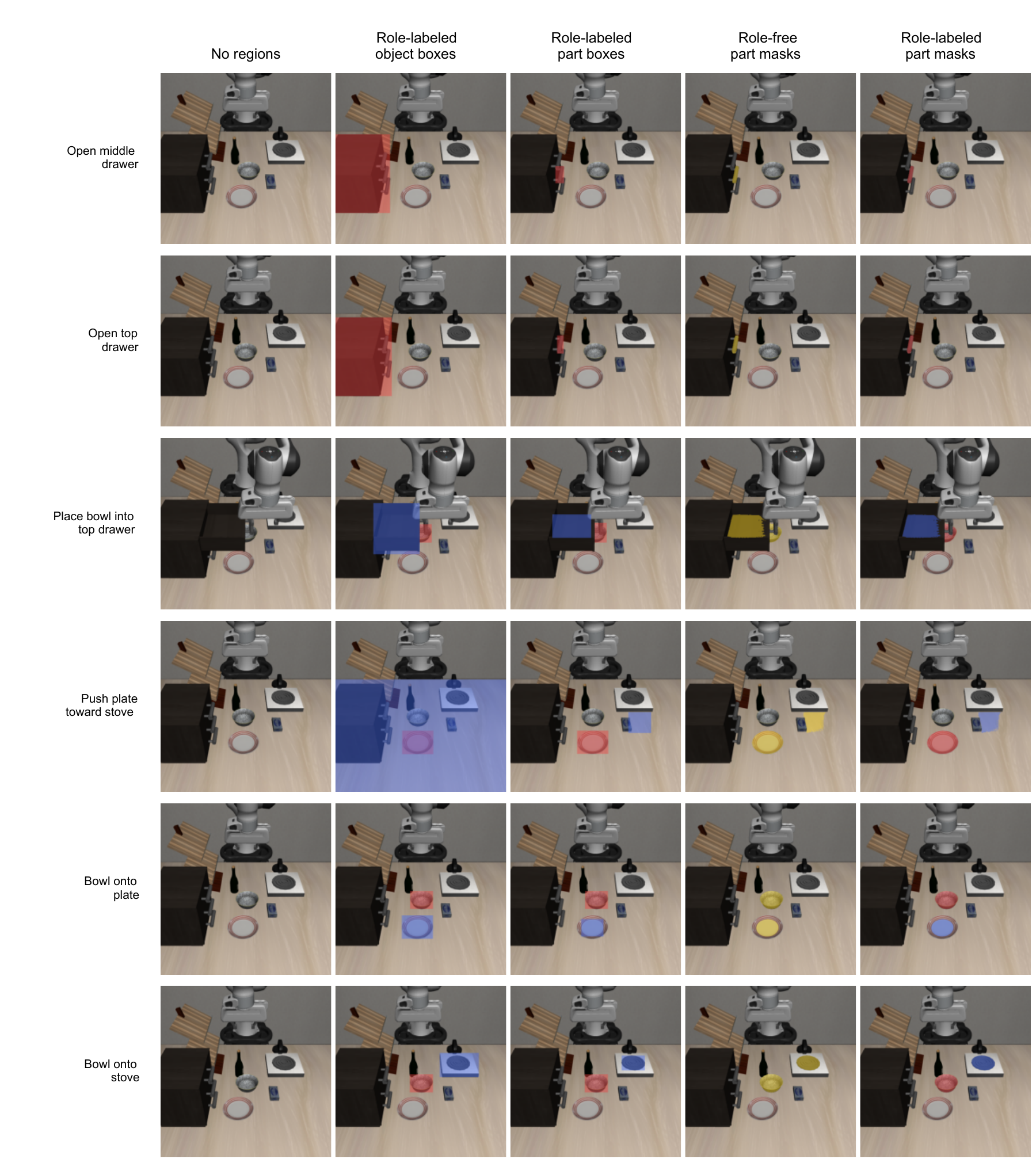}
\caption{\textbf{Region conditions across manipulation task steps.} Columns show no regions, role-labeled object boxes, role-labeled part boxes, role-free part masks, and role-labeled part masks for the same initial observation. Drawer opening and bowl placement are shown separately for the two-step task. Boxes are rasterized for the policy; overlays are visualizations, not RGB inputs. Continued on the next page.}
\label{fig:supp-act-inputs}
\end{figure}
\clearpage
\begin{figure}[H]
\ContinuedFloat
\centering
\includegraphics[width=0.88\linewidth]{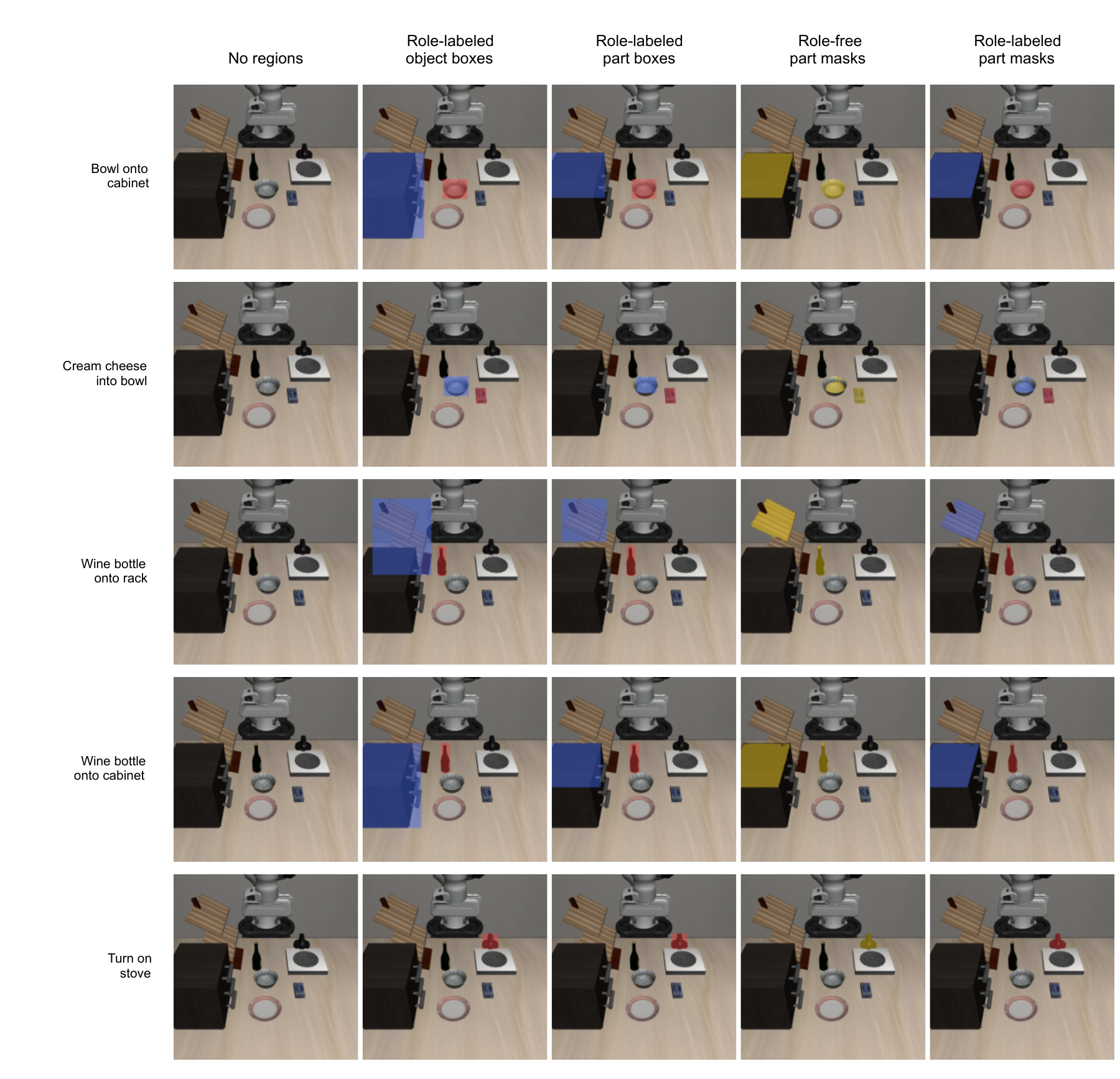}
\caption{\textbf{Region conditions across manipulation task steps (continued).} The remaining task steps use the same five input conditions. ``Role-free'' denotes anonymous independent part masks without fixed role-slot identities; the individual regions are retained. Colors illustrate supplied regions and are not added to the policy's RGB inputs.}
\end{figure}

\label{sec:supp-act-conditions}
To examine alternative region representations for policy conditioning, we compare five inputs using a common policy architecture and training protocol. All conditions receive current and initial RGB observations, proprioception, action text, and a verb indicator. We vary the regions on the initial image: no regions, role-labeled full-object boxes, role-labeled part boxes, anonymous part masks, and role-labeled part masks. Boxes are rasterized as rectangular masks and processed by the same region-token pathway. Full-object boxes are manually supplied for all 486 training and validation anchors; part boxes are derived from the part masks.

For anonymous masks, regions are canonicalized by content and permuted with a sample-specific seed; presence indicators follow the same permutation. The permutation remains fixed throughout a step. This removes explicit mask-to-role slot assignment while retaining the separate regions and action text. Role-labeled masks instead occupy fixed O/I/D slots. The no-region condition masks out region tokens and sets their presence indicators to zero while retaining global initial-image features. \Cref{fig:supp-act-inputs} and its continuation illustrate these inputs across all task-step types. The drawer-opening examples show how an object box can cover the cabinet while a part region singles out its handle. Role-free and role-labeled masks retain the same region geometry while differing in role assignment.
\clearpage
\subsection{Policy and training}
\label{sec:supp-act-architecture}
The ACT-derived policy~\cite{zhao2023learning} is illustrated in \cref{fig:supp-act-architecture}. A shared ImageNet-initialized ResNet18~\cite{he2016deep} encodes current and initial RGB images, resized from $512\times512$ to $224\times224$. Each final $7\times7$ feature grid is projected to width 256, giving 98 global visual tokens with spatial and view encodings. The region branch uses the initial image's $28\times28$ intermediate features, projected to the same width. Mask coverage is area-pooled to this grid; its logarithm biases attention, with zero-coverage positions masked. Four learned queries per slot attend to the covered image features, producing twelve region tokens that are appended to the global visual tokens at the ACT encoder input. Absent slots are zeroed and padding-masked.

A state token combines 16-dimensional proprioception, a six-dimensional verb indicator, three slot-presence indicators, and a 128-dimensional projection of a frozen 384-dimensional MiniLM~\cite{wang2020minilm} text embedding. The action transformer has width 256, eight heads, feed-forward width 1,024, and four encoder layers. The implementation instantiates four decoder layers, with the action head reading the first intermediate decoder-layer output. The CVAE posterior encoder has four layers and a 32-dimensional latent; inference fixes the latent to zero. Each prediction is a chunk of sixteen seven-dimensional \texttt{OSC\_POSE} actions. All conditions use this implementation.

\Needspace{0.30\textheight}
\paragraph{Training.}
\label{sec:supp-act-training}
Each condition is trained independently from the same initialization, with seed 42 and 30,000 optimizer updates, approximately 9.76 epochs. AdamW uses learning rates $10^{-4}$ for the policy and $10^{-5}$ for the visual backbone, weight decay $10^{-4}$, and gradient clipping at 1.0. A microbatch of four with four accumulation steps gives a nominal effective batch of sixteen. Training uses BF16 and trainable backbone convolutions with frozen batch normalization. The objective combines masked normalized-action L1 with CVAE KL at weight 1.0; the text encoder and unused stop head are frozen.

Checkpoints are selected by the lowest validation action L1 with the inference latent fixed to zero (\cref{tab:supp-act-training}). Execution success is not used for checkpoint selection.
\begin{table}[H]
\centering
\caption{\textbf{Validation-selected policy checkpoints under a shared 30,000-update budget.} Epochs are one-based; L1 is evaluated with the inference latent fixed to zero.}
\label{tab:supp-act-training}
\small
\setlength{\tabcolsep}{4pt}
\begin{tabularx}{\linewidth}{@{}Yrrr@{}}
\toprule
Region condition & Epoch & Update & Val. L1 \\
\midrule
No regions & 3 & 9,222 & 0.30987 \\
Role-labeled object boxes & 2 & 6,148 & 0.30881 \\
Role-labeled part boxes & 2 & 6,148 & 0.30663 \\
Anonymous part masks & 2 & 6,148 & 0.31184 \\
Role-labeled part masks & 3 & 9,222 & 0.30818 \\
\bottomrule
\end{tabularx}
\end{table}

\Needspace{0.40\textheight}
\subsection{Evaluation protocol}
\label{sec:supp-act-evaluation}
During each step, initial RGB, action, and supplied regions remain fixed, while current RGB and proprioception update during execution. At 20\,Hz, the policy predicts sixteen actions, executes the first four, and replans, up to 200 actions per step. Completion is checked after every action with simulator predicates: the intermediate drawer-opening step uses its specified BDDL open predicate, and other steps use the corresponding task-success predicate.

Both steps of the two-step task are restored separately to their demonstration states. The metric is therefore micro-averaged step success. All cases begin with a false task-success flag and execute nonzero actions. Every condition uses paired initial states and all 51 cases; no runtime errors or pending cases remain.

\label{sec:supp-act-scope}
Policy and perception checkpoint selection, together with architecture development, use the reported validation collection.

\Needspace{6\baselineskip}
\subsection{Additional results}
\label{sec:supp-act-results}
\Cref{tab:supp-act-conditions} supplements the main comparison in \cref{tab:downstream-utility} with all five GT region inputs. Per-task and paired outcomes are given in \cref{tab:supp-act-tasks,tab:supp-act-paired}.
\begin{table}[H]
\centering
\caption{\textbf{Ground-truth region inputs with independently trained instances of the region-token policy.} Each condition evaluates the same 51 validation steps.}
\label{tab:supp-act-conditions}
\small
\setlength{\tabcolsep}{4pt}
\begin{tabularx}{\linewidth}{@{}Yrr@{}}
\toprule
Region condition & Successful steps & Success (\%) \\
\midrule
No regions & 41/51 & 80.4 \\
Role-labeled object boxes & 27/51 & 52.9 \\
Role-labeled part boxes & 35/51 & 68.6 \\
Anonymous part masks & 38/51 & 74.5 \\
Role-labeled part masks & \textbf{45/51} & \textbf{88.2} \\
\bottomrule
\end{tabularx}
\end{table}

\paragraph{Predicted masks.}
\label{sec:supp-act-predicted}
For the perception adaptation described in \cref{sec:experiments-4}, language LoRA uses learning rate $2\times10^{-6}$ and the text-to-mask projection and SAM2 mask decoder use $10^{-5}$; other components are frozen. Training uses eight H100 GPUs, global batch 32, and at most 20 epochs. Epoch 16 is selected on the 51 validation image--action pairs.

For execution, perception receives each step's initial RGB image and given action. Predicted regions condition the fixed GT-trained executor, without GT fallback for invalid outputs. The rollout cache yields 94.27\% cIoU and 91.02\% mean IoU over GT-positive role instances, with 100\% format validity and exact role-presence accuracy. This instance-pooled mean differs from macro-averaging role-wise NE-mIoU.

The two changed GT/predicted execution outcomes reported in \cref{tab:downstream-utility} occur in the independent bowl-placement step of one demonstration and placement onto the stove in another.

\Needspace{0.70\textheight}
\paragraph{Region integration.}
\label{sec:supp-act-fusion}
To examine sensitivity to the region-integration design, \cref{tab:supp-act-fusion} compares early feature addition with region-token conditioning. Feature addition encodes the regions and adds their features to the step-start image branch at $28\times28$ resolution, before the later visual layers. Region-token conditioning pools step-start image features under the supplied regions and appends the resulting tokens to the ACT encoder input. With role-labeled part masks, region-token conditioning achieves 45/51 successful steps compared with 41/51 for feature addition. Relative to each architecture's no-region condition, these masks add four successes with region-token conditioning and reduce the count by one with feature addition. This comparison shows how the region-integration design affects the utility of role-labeled masks; the main-text study uses region-token conditioning.

\begin{table}[H]
\centering
\caption{\textbf{Per-task step successes with region-token fusion.} O-box and P-box denote role-labeled object and part boxes; anonymous and role masks use part regions. The two-step task aggregates separately initialized steps.}
\label{tab:supp-act-tasks}
\small
\setlength{\tabcolsep}{4pt}
\begin{tabularx}{\linewidth}{@{}Yrrrrr@{}}
\toprule
Task & None & O-box & P-box & Anon. & Role \\
\midrule
Open the middle drawer of the cabinet & 2/5 & 0/5 & 1/5 & 0/5 & 5/5 \\
Open the top drawer and put the bowl inside & 8/8 & 7/8 & 7/8 & 6/8 & 7/8 \\
Push the plate to the front of the stove & 4/4 & 2/4 & 0/4 & 2/4 & 4/4 \\
Put the bowl on the plate & 5/5 & 1/5 & 2/5 & 4/5 & 5/5 \\
Put the bowl on the stove & 5/5 & 1/5 & 4/5 & 4/5 & 5/5 \\
Put the bowl on top of the cabinet & 5/5 & 5/5 & 4/5 & 5/5 & 5/5 \\
Put the cream cheese in the bowl & 3/5 & 0/5 & 3/5 & 3/5 & 3/5 \\
Put the wine bottle on the rack & 2/4 & 2/4 & 4/4 & 4/4 & 2/4 \\
Put the wine bottle on top of the cabinet & 4/5 & 4/5 & 5/5 & 5/5 & 5/5 \\
Turn on the stove & 3/5 & 5/5 & 5/5 & 5/5 & 4/5 \\
\bottomrule
\end{tabularx}
\par\medskip
\begin{minipage}[t]{0.48\linewidth}
\centering
\caption{\textbf{Paired outcomes comparing role-labeled part masks with each alternative.} Role only and alternative only count successes unique to that condition.}
\label{tab:supp-act-paired}
\small
\setlength{\tabcolsep}{4pt}
\begin{tabularx}{\linewidth}{@{}Yrrrr@{}}
\toprule
Alternative & \shortstack{Role\\only} & \shortstack{Alternative\\only} & \shortstack{Both\\succeed} & \shortstack{Both\\fail} \\
\midrule
No regions & 6 & 2 & 39 & 4 \\
Role-labeled object boxes & 21 & 3 & 24 & 3 \\
Role-labeled part boxes & 13 & 3 & 32 & 3 \\
Anonymous part masks & 10 & 3 & 35 & 3 \\
\bottomrule
\end{tabularx}
\end{minipage}\hfill
\begin{minipage}[t]{0.48\linewidth}
\centering
\caption{\textbf{Step successes under two region-integration mechanisms.} All conditions use training seed 42 and evaluation seed 0 on the same 51 steps.}
\label{tab:supp-act-fusion}
\small
\setlength{\tabcolsep}{4pt}
\begin{tabularx}{\linewidth}{@{}Yrr@{}}
\toprule
Region condition & Feature addition & Region tokens \\
\midrule
No regions & 42/51 & 41/51 \\
Role-labeled object boxes & 42/51 & 27/51 \\
Role-labeled part boxes & 39/51 & 35/51 \\
Anonymous part masks & 39/51 & 38/51 \\
Role-labeled part masks & 41/51 & 45/51 \\
\bottomrule
\end{tabularx}
\end{minipage}
\end{table}

\clearpage
\subsection{Real-robot protocol}
\label{sec:appendix-real-robot-protocol}
The real-robot comparison in \cref{sec:real-robot} uses the same execution interface and budgets across conditions. The reference plans below specify task scoring and role combinations.

The three tasks exercise different O/I/D combinations, as illustrated by the reference steps below. Unlisted roles are absent, and robot grippers are not instruments under our schema.

\paragraph{Place the cabbage into the pot, then put the lid on the pot.}
The reference steps are:
\begin{itemize}
    \item Place the cabbage into the pot. \textbf{O+D}: O is the cabbage; D is the receiving region of the pot.
    \item Put the lid on the pot. \textbf{O+D}: O is the lid; D is the receiving region of the pot.
\end{itemize}
Each completed action contributes 0.5. Full success requires the cabbage to be inside the covered pot.

\paragraph{Put the pen into the drawer, then close the drawer.}
Exactly one pen is present. The reference steps are:
\begin{itemize}
    \item Place the pen into the drawer. \textbf{O+D}: O is the pen; D is the drawer interior.
    \item Close the drawer. \textbf{O}: O is the drawer; I and D are absent.
\end{itemize}
Pen placement and drawer closure each contribute 0.5. Full success requires the pen to be inside the closed drawer.

\paragraph{Sweep the block into the dustpan.}
A reference plan may include:
\begin{itemize}
    \item Position the dustpan beside the block (optional). \textbf{O+D}: O is the dustpan; D is its intended placement region beside the block.
    \item Sweep the block into the dustpan with the brush. \textbf{O+I+D}: O is the block; I is the brush's working region; D is the receiving surface of the dustpan.
\end{itemize}
To keep scoring independent of plan length, sweeping receives 1 if the block ends in the dustpan and 0 otherwise. Positioning the dustpan receives no separate credit.

\smallskip
\noindent
These reference steps may differ from the generated plans. Mean task-completion scores are reported as percentages, and a rollout scoring 1 counts as a full-task success.

Each rollout starts with empty execution memory and permits up to 40 control rounds. A control round is one iteration of the execution loop, normally containing one executor decision based on current observations, the overall task, the current step, and memory. The decision requests measurement, motion, or step/plan management. Measurement results inform subsequent decisions, while action sequences execute without intermediate model decisions. Invalid decisions and rejected actions also consume a round.

A separate budget of 60 logical model requests is shared by Astra and Luna over the entire rollout. It includes initial planning and replanning, mask review, executor decisions, and further requests for correction or refreshed role references. A control round can therefore consume multiple model calls. Network retries of the same logical request and non-model operations such as depth processing, inverse kinematics, and robot control do not add model calls.

\subsection{Real-robot resource use}
\label{sec:appendix-real-robot-resources}
With Luna execution, execution calls increase from 21.21 to 27.07, and mean runtime increases from 321.3 to 446.7 seconds (\cref{tab:supp-real-robot-cost}). With Astra execution, execution calls remain close at 18.86 and 18.73, and runtime increases from 291.8 to 312.9 seconds.

\begin{table}[H]
\centering
\caption{\textbf{Real-robot resource use per rollout.} Values are averages; $N$ denotes the number of rollouts summarized in each row. Planning uses Astra in all conditions. Execution calls use the indicated executor. Tokens and wall time include the representation branch when used.}
\label{tab:supp-real-robot-cost}
\small
\setlength{\tabcolsep}{4pt}
\begin{tabularx}{\linewidth}{@{}YYrrrrr@{}}
\toprule
 & & & \multicolumn{2}{c}{Model calls} & & \\
\cmidrule(lr){4-5}
Executor & Role conditioning & $N$ & Planning & Execution & Tokens & Time (s) \\
\midrule
Luna & Without & 29 & 1.00 & 21.21 & 133,832 & 321.3 \\
Luna & With & 29 & 1.00 & 27.07 & 182,832 & 446.7 \\
Astra & Without & 28 & 1.00 & 18.86 & 126,307 & 291.8 \\
Astra & With & 30 & 1.00 & 18.73 & 132,836 & 312.9 \\
\bottomrule
\end{tabularx}
\end{table}

Luna's published Standard API input and output token rates are each one hundredth of Astra's~\cite{openai_api_pricing}, motivating repeated control decisions with a lower-cost executor.